\documentclass{article}
\usepackage[utf8]{inputenc}
\usepackage{amsmath,amsthm,setspace,amssymb,esint,fancyhdr,lastpage,extramarks,chngpage,enumerate}
\usepackage{comment}

\usepackage[backend=biber, style=numeric, sorting=nty]{biblatex}
\usepackage{geometry}[margin=1in]   
\usepackage{amsfonts}
\usepackage{amsmath}

\usepackage{url}
\usepackage{graphicx}
\usepackage{subcaption}
\usepackage{authblk}
\usepackage[font={small,it}]{caption}
\usepackage{threeparttable}
\usepackage{caption}
\usepackage{hyperref}
\usepackage[symbol]{footmisc}

\usepackage{tabulary}
\usepackage{booktabs}
\usepackage[table]{xcolor}

\title{\texttt{Weak-PDE-LEARN}: A Weak Form Based Approach to Discovering PDEs From Noisy, Limited Data}
\author[a]{Robert Stephany\thanks{Corresponding Author. 

Email address: \href{mailto:rrs254@cornell.edu}{\texttt{rrs254@cornell.edu}}}$^{,}$}
\author[a,b]{Christopher Earls}
\affil[a]{\it Center for Applied Mathematics, 657 Frank H.T. Rhodes Hall, Cornell University, Ithaca, NY 14853, United States}
\affil[b]{\it School of Civil \& Environmental Engineering, Cornell University, Ithaca, NY 14853, United States}
\date{}

\begin{document}

\maketitle

\begin{center} {\bf Abstract} \end{center}
We introduce \texttt{Weak-PDE-LEARN}, a Partial Differential Equation (PDE) discovery algorithm that can identify non-linear PDEs from noisy, limited measurements of their solutions.
\texttt{Weak-PDE-LEARN} uses an adaptive loss function based on weak forms to train a neural network, $U$, to approximate the PDE solution while simultaneously identifying the governing PDE.
This approach yields an algorithm that is robust to noise and can discover a range of PDEs directly from noisy, limited measurements of their solutions.
We demonstrate the efficacy of \texttt{Weak-PDE-LEARN} by learning several benchmark PDEs.

\bigskip 


\section{Introduction}
\label{Sec:Introduction}

One of the central goals of science is to discover predictive models to describe the world around us; to distill natural phenomena into mathematical formulas that we can analyze.
Historically, the scientific community discovered these models by observing natural phenomena and then attempting to distill the \textit{first principles} that govern the observed behavior.
Scientists seek a model whose predictions match real-world observations. 
This approach has yielded predictive models in a myriad of fields, from fluid mechanics to general relativity.
Unfortunately, many important systems, particularly those in the biological sciences \cite{abreu2019mortality}, \cite{amor2020transient}, lack predictive models.

\bigskip

Machine learning offers an intriguing alternative to the first-principles approach.
Notably, pattern recognition is one of the core strengths of machine learning (ML), making ML a natural choice for discovering scientific models describing complex observational data.
Correspondingly, in recent years, several pioneering works have demonstrated deep learning's ability to aid scientific research.
Researchers have employed machine learning models to discover Green's functions \cite{boulle2022data} \cite{gin2021deepgreen}, symmetries \cite{boulle2022data}, Hamiltonian's \cite{greydanus2019hamiltonian}, Dynamical Systems \cite{brunton2016discovering}, Delay Dynamical Systems \cite{oprea2023learning}, and invariant quantities \cite{schmidt2009distilling} directly from scientific data.
Additionally, the literature is replete with many other important contributions.

\bigskip

In this paper, we focus on a subset of these earlier efforts: discovering governing Partial Differential Equations (PDEs) from measurements of their solutions, also known as PDE discovery.
We discuss existing approaches to PDE discovery in section \ref{Sec:Related_Work}.
In general, PDE discovery seeks to identify the Partial Differential Equations that govern natural phenomena directly from measurements of those phenomena.
PDE discovery algorithms can help scientists discover predictive models for natural phenomena that have been difficult to model using the first principles approach.

\bigskip

Crucially, any algorithm that hopes to contribute to scientific discovery must be able to work with scientific measurements.
Collecting high-quality scientific data can be expensive and arduous.
Therefore, scientific measurements tend to be limited (there are few measurements) and noisy.
Thus, to be practical, a PDE discovery algorithm must be able to identify a PDE from noisy, limited measurements of its solutions.

\bigskip

{\bf Contributions:} This paper builds upon our earlier work on the \texttt{PDE-LEARN} algorithm. 
In this paper, we introduce a novel PDE discovery algorithm, \texttt{weak-PDE-LEARN}, which discovers PDEs directly from measurements of their solutions.
\texttt{Weak-PDE-LEARN} extends earlier work by integrating the unknown PDE against a collection of specially selected \emph{weight functions}.
This approach allows us to learn the PDE without directly approximating the classical partial derivatives of its solutions; rather, we employ generalized partial derivatives \cite{reddy2013introductory}.
We also introduce an algorithm that uses the available measurements to adaptively select the weight functions.
This adaptive approach helps accelerate PDE identification. 
We demonstrate \texttt{Weak-PDE-LEARN}'s efficacy by discovering a variety of benchmark PDEs from limited, noisy measurements of their solutions.

\bigskip 

{\bf Outline:} We have organized the rest of the paper as follows:
In section \ref{Sec:Related_Work}, we survey algorithms for discovering scientific laws from data, emphasizing algorithms for PDE discovery.
Next, in section \ref{Sec:Problem}, we formally describe the problem at hand, including our assumptions about the form of the PDEs we aim to discover.
Then, In section \ref{Sec:Method}, we describe the \texttt{Weak-PDE-LEARN} algorithm. 
In section \ref{Sec:Experiments}, we use \texttt{Weak-PDE-LEARN} to discover several benchmark PDEs from noisy, limited measurements of their solutions. 
Section \ref{Sec:Discussion} discusses the rationale behind our algorithm and some of its limitations. 
Finally, we give concluding remarks in section \ref{Sec:Conclusion}.

\section{Problem Statement}
\label{Sec:Problem}

\texttt{weak-PDE-LEARN} identifies a function, $u$, and a PDE that it satisfies by studying noisy measurements of $u$. 
In this section, we make the preceding sentence more precise.
First, we define the system response function, problem domain, and noise level.
We then state our assumptions about the form of the PDE that the system response function, $u$, satisfies.

\subsection{System Response Function, Noise}
\label{SubSec:System_Response}

Let $T > 0$ and $\Omega \subseteq \mathbb{R}^{d}$ be a compact, connected set with a Lipschitz boundary. 
We assume there is a function, $u : [0, T] \times \Omega \to \mathbb{R}$, called the \emph{system response function}.
We refer to $\Omega$, $[0, T]$, and $[0, T] \times \Omega$ as the \emph{spatial}, \emph{temporal}, and \emph{problem domains}, respectively. 
This paper primarily focuses on the case when $d = 1$ (one spatial variable), primarily because it is possible to visualize the solutions to such problems, though \texttt{weak-PDE-LEARN} works for an arbitrary $d$.

\bigskip

We assume we have a finite set of noisy measurements, $\{ \tilde{u}(t_i, X_i) \}_{i = 1}^{N_{Data}}$, of the system response function.
Throughout this paper, we will refer to $\{ \tilde{u}(t_i, X_i) \}_{i = 1}^{N_{Data}}$ as the \emph{noisy data set}. 
Likewise, we will refer to $\{ u(t_i, X_i) \}_{i = 1}^{N_{Data}}$ as the \emph{noise-free data set}.

\bigskip

In general, since the measurements are noisy, we have $\tilde{u}(t_i, X_i) \neq u(t_i, X_i)$.
We define the \emph{noise} at the $i$th data point, $t_i, X_i$, as the difference $\tilde{u}(t_i, X_i) - u(t_i, X_i)$.
 We assume the noise represents samples from a set of i.i.d. Gaussian random variables with mean zero.
That is, we assume there exists some $\sigma > 0$ such that for each $i \in \{ 1, 2, \ldots, N_{Data} \}$,

\begin{equation} \big( \tilde{u}(t_i, X_i) - u(t_i, X_i) \big) \sim N\left(0, \sigma^2 \right). \label{Eq:Noise}\end{equation}

Finally, we define the \emph{noise level} in a data set as the ratio of $\sigma$ to the standard deviation of the measurements in the corresponding noise-free data set.
That is,

\begin{equation}\text{noise level} = \frac{\sigma}{SD \Big( u\big(t_1, X_1\big), \ldots, u\big(t_{N_{Data}}, X_{N_{Data}} \big) \Big) }. \label{Eq:Noise_Level} \end{equation}

\subsection{The Hidden PDE}
\label{SubSec:Hidden_PDE}

We assume the system response function, $u$, satisfies a partial differential equation (PDE) of the form 

\begin{equation} D^{\alpha(0)} F_0\big(u(t, X) \big) = \sum_{m = 1}^{M} c_m D^{\alpha(m)} F_m\big( u(t, X) \big) \qquad\qquad (t, X) \in \Omega. \label{Eq:PDE} \end{equation}

We refer to equation \eqref{Eq:PDE} as the \emph{hidden PDE}.
Each $D^{\alpha(m)}$ is a known partial derivative operator characterized by the multi-index $\alpha(m)$. 
For example, if $\alpha(m) = (2, 3)$, then 

$$D^{\alpha(m)} F_m = \frac{\partial^2}{\partial t^2} \left( \frac{\partial^3}{\partial x^3} F_m \right).$$

For clarity, we often write the derivative operator $D^{(\alpha(1), \alpha(2))}$ as 
$$D^{(\alpha(1), \alpha(2))} = D_t^{\alpha(1)} D_x^{\alpha(2)}.$$
Note that if one of the indices is $1$, we usually write $D_t$ and $D_x$ in place of of $D_t^1$ and $D_x^1$, respectively.

\bigskip 

In section \ref{Sec:Method}, we show how to exploit equation \eqref{Eq:PDE} to derive a method to learn the hidden PDE without evaluating the partial derivatives of $u$.

\bigskip

Each function, $F_m$, is a known function of $u$, while the coefficients, $c_m$, are unknown.
We will refer to $D^{\alpha(1)} , \ldots, D^{\alpha(M)}$ and $D^{\alpha(1)} F_1, \ldots, D^{\alpha(M)} F_M$ as the \emph{derivative operators} and \emph{library terms}, respectively.
We refer to $D^{\alpha(0)} F_0$ as the \emph{left-hand side term} (or \emph{LHS term} for short) and the collection $\{ D^{\alpha(1)} F_1, \ldots, D^{\alpha(M)} F_M \}$ as the \emph{right-hand side terms} (or \emph{RHS Terms} for short). 

\bigskip 

Our algorithm can, in general, work with any functions, $F_1, \ldots, F_M$.
Throughout this paper, however, we will restrict our attention to the case when each $F_m(u)$ is a monomial of $u$. 
That is, for each $m \in \{ 1, 2, \ldots, M \}$, there is some $k_m \in \mathbb{N} \cup \{ 0 \}$ such that

$$F_m(u) = u^{k_m}.$$

\subsection{Coefficient sparsity}
\label{SubSec:Sparsity}

In general, we do not know the exact form of the hidden PDE \emph{a priori}.
The goal, after all, is to discover a hidden, governing PDE using noisy measurements of its solution. 
We assume the right-hand side of the hidden PDE is a sparse linear combination of the library terms. 
More concretely, we assume most of the right-hand side terms have a coefficient of $0$ and, therefore, do not contribute to $u$'s dynamics.

\bigskip

Thus, equation \ref{Eq:PDE} merely specifies the general form of the hidden PDE: we assume the terms in the hidden PDE belong to a larger set of library terms.
In general, the larger the library, the larger the set of PDEs we can express using those terms.
With that said, increasing the size of the library increases the number of terms we need to eliminate to identify the hidden PDE.
Throughout this paper, we use a set of library terms much larger than the set that actually contributes to the underlying dynamics, as this situation more closely resembles what might occur in practice (when the hidden PDE is unknown).

\subsection{Objective}
\label{SubSec:Goals}

The goal of \texttt{weak-PDE-LEARN} is, to use the noisy data set, $\left\{ \tilde{u}(t_i, X_i) \right\}_{i = 1}^{N_{Data}}$, and the library terms, $\{ D^{\alpha(m)} F_m \}_{m = 1}^{M}$, to learn the sparse coefficients $c_1, \ldots, c_M$.

\bigskip 

Table \ref{Table:Problem:Notation} lists the notation we introduced in the forgoing section.

\bigskip

\begin{table}[hbt]
    \centering 
    \rowcolors{2}{white}{cyan!10}
    
    \begin{tabulary}{1.0\linewidth}{p{3.1cm}L}
        \toprule[0.3ex]
        \textbf{Notation} & \textbf{Meaning} \\
        \midrule[0.1ex]

        $d$ & The number of spatial dimensions in the spatial problem domain. \\
        \addlinespace[0.4em]
        $[0, T] \times \Omega \subseteq \mathbb{R} \times \mathbb{R}^d$ & The problem domain. $[0, T] \subseteq \mathbb{R}$ and $\Omega \subseteq \mathbb{R}^d$ refer to the temporal and spatial parts, respectively. \\
        \addlinespace[0.4em]

        $u : [0, T] \times \Omega_i \to \mathbb{R}$ & The system response function. \\
        \addlinespace[0.4em]

        $\tilde{u}(t, X)$ & A noisy measurement of $u$ at $(t, X) \in (0, T] \times \Omega$. \\
        \addlinespace[0.4em]
        Noisy data set & The set $\{ \tilde{u}(t_i, X_i) \}_{i = 1}^{N_{Data}}$ of measurements. This is what we actually train \texttt{weak-PDE-LEARN} on. \\
        \addlinespace[0.4em]
        Noise-free data set & The set $\{ u(t_i, X_i) \}_{i = 1}^{N_{Data}}$. We do not assume knowledge of this set. \\
        \addlinespace[0.4em]
        Noise level & The ratio of the standard deviation of the noise to that of the noise-free data set. See equation \eqref{Eq:Noise}. \\
        \addlinespace[0.4em]

        $M$ & The number of right-hand side terms in the hidden PDE, equation \eqref{Eq:PDE}. \\
        \addlinespace[0.4em]
        $D^{\alpha(0)}, \ldots, D^{\alpha(M)}$ & The derivative operators in the hidden PDE, equation \eqref{Eq:PDE}. \\
        \addlinespace[0.4em]
        $F_0, \ldots, F_m$ & The functions in the hidden PDE terms. See equation \eqref{Eq:PDE}. \\
        \addlinespace[0.4em]
        LHS term & $D^{\alpha(0)} F_0.$ \\
        \addlinespace[0.4em]
        RHS terms & $\{ D^{\alpha(1)} F_1, \ldots, D^{\alpha(M)} F_M.$ \\
        \addlinespace[0.4em]
        $c_1, \ldots, c_K$ & The coefficients of the RHS terms terms $f_1, \ldots, f_K$ in equation \ref{Eq:PDE}. \\
        \bottomrule[0.3ex]
    \end{tabulary}
    
    \caption{The notation and terminology of section (\ref{Sec:Problem})} 
    \label{Table:Problem:Notation}
\end{table}

\section{Related Work}
\label{Sec:Related_Work}

\bigskip 

The modern, machine-learning approach to the automatic discovery of scientific laws from data began in the late 2000s with \cite{bongard2007automated} and \cite{schmidt2009distilling}.
The former, \cite{bongard2007automated}, uses genetic algorithms to discover the governing equation for a dynamical system using noisy measurements of one of its solutions.
The latter, \cite{schmidt2009distilling}, uses a similar approach to discover conservation laws from scientific data. 
Both use noisy measurements of the system response function to learn properties of the underlying dynamics. 

\bigskip 

One of the most important contributions was the Sparse Identification of Nonlinear DYnamics (\texttt{SINDY}) algorithm \cite{brunton2016discovering}.
\texttt{SINDY} assumes the user has noisy measurements of the system response function evaluated on a regular grid in the problem domain.
It uses finite difference (or other standard numerical differentiation techniques) to approximate derivatives of the system response function.
\texttt{SINDY} then uses these values to set up a linear system of equations whose sparse solution characterizes a dynamical system that the system response function satisfies.
It finds the spare solution using the \texttt{ST-Ridge} algorithm, which solves a sequence of progressively sparser ridge regression ($L^2$ penalized least-squares) problems. 

\bigskip

Shortly after the introduction of SINDY, Rudy et al. proposed \texttt{PDE-FIND}, a modification of \texttt{SINDY} that can identify PDEs from measurements of one of its solutions \cite{rudy2017data}. 
\texttt{PDE-FIND} uses similar assumptions to the ones we discuss in section \ref{Sec:Problem}.
Additionally, like \texttt{SINDY}, \texttt{PDE-FIND} assumes the user has measurements of the system response function evaluated on a regular grid in the problem domain.
This restriction allows \texttt{PDE-FIND} to evaluate the partial derivatives of the system response function using standard numerical differentiation techniques, such as finite differences.
Using these values, \texttt{PDE-FIND} evaluates the library of terms at the data points, which engenders a linear system for the coefficients of the hidden PDE, $c_1, \ldots, c_K$.
\texttt{PDE-FIND} then identifies the coefficients using \texttt{ST-Ridge}.
\texttt{PDE-FIND} can identify many benchmark PDEs directly from noisy measurements.
With that said, \texttt{PDE-FIND} does have some limitations.
In particular, since numerical differentiation tends to amplify noise, noise considerably degrades \texttt{PDE-FIND}'s abilities.
Further, requiring the measurements to occur on a regular grid may be impractical.

\bigskip

Two other early examples of PDE discovery algorithms are \cite{schaeffer2017learning} and \cite{berg2019data}.
The former proposes an approach similar to \texttt{PDE-FIND} but approximates the derivatives using spectral methods.
This change makes their algorithm quite resilient to noise. 
Like \texttt{PDE-FIND}, however, their approach assumes the user has measurements of the system response function evaluated on a regular grid on the problem domain.
The latter, \cite{berg2019data}, trains a neural network, $U : (0, T] \times \Omega \to \mathbb{R}$, to approximate the system response function.
After training, it uses a sparse regression algorithm to identify a PDE that the system response function satisfies.
Using a neural network to interpolate the measurements of the system response function allows the data points to be distributed arbitrarily throughout the problem domain.

\bigskip 

There have also been successful attempts to use a weak-formulation approach to discovering scientific laws.
For PDE discovery, two notable examples are \cite{gurevich2019robust} and \cite{messenger2021weak}.
Both learn the hidden PDE using a similar approach to the one introduced in section \ref{SubSec:Method:Weight}.
Both papers use weight functions that are defined to be a polynomial on rectangle in the problem domain and zero everywhere else.
These weight functions are not infinitely differentiable, but they are smooth enough to use integration by parts to offload the derivatives in library terms to the weight function (a technique we discuss in detail in section \ref{SubSubSec:Method:Weak_Form}). 
Both approaches demonstrate impressive robustness to noise. 
In particular, \cite{gurevich2019robust} is able to learn the Kuramoto-Sivashinsky equation, a PDE that many PDE discovery algorithms struggle with, even in the presence of considerable noise.
Further, \cite{messenger2022asymptotic} recently explored these approaches from a theoretical perspective and established several key results about them.
They showed that these approaches yield an estimator that, given enough data, can identify a wide array of PDEs from noisy measurements.

\bigskip

More recently, the authors of this paper introduced \texttt{PDE-LEARN} \cite{stephany2022LEARN}, which has several features in common with the algorithm we preset in this paper.
\texttt{PDE-LEARN} uses a neural network, $U$, and a trainable vector, $\xi \in \mathbb{R}^M$, to approximate the system response function and the coefficients $c_1, \ldots, c_M$ in the right-hand side of the hidden PDE, respectively.
\texttt{PDE-LEARN} trains $U$ and $\xi$ using a three-part loss function.
One part of the loss function is the \emph{collocation loss} (which takes the place of the \emph{weak form loss} we present in section \ref{Sec:Method}).
The collocation loss requires that $U$ approximately satisfies the PDE encoded in $\xi$, thereby coupling the training of $\xi$ and $U$. 
To enforce this, \texttt{PDE-LEARN} uses automatic differentiation \cite{baydin2018automatic} to compute the partial derivatives of $U$ at a randomly sampled collection of \emph{collocation points} in the problem domain.
This approach allows \texttt{PDE-LEARN} to directly evaluate the left and right-hand sides of the hidden PDE, equation \eqref{Eq:PDE}.
\texttt{PDE-LEARN} then computes the mean square error between the left and right-hand side of the hidden PDE evaluated at these points (with the components of $\xi$ replacing those of $c$, and $U$ replacing $u$).
The resulting value is collocation loss.
\texttt{PDE-LEARN} also uses an adaptive process to place additional collocation points in regions of the problem domain where the left and right-hand sides of the hidden PDE differ the most. 
\texttt{PDE-LEARN} successfully identifies several benchmark PDEs from noisy, limited measurements of the system response function.

\bigskip

Finally, some other notable examples of PDE discovery algorithms include \cite{stephany2022READ}, \cite{atkinson2019data}, and  \cite{bonneville2021bayesian}.
\cite{stephany2022READ} introduced an algorithm called \texttt{PDE-READ}.
\texttt{PDE-READ} uses two neural networks: The first, $U : (0, T] \times \Omega \to \mathbb{R}$, learns the system response function, and the second, $N$, learns an abstract representation of the right-hand side of equation \ref{Eq:PDE}. 
That approach is similar to Raissi’s \emph{deep hidden physics models} algorithm \cite{raissi2018deep}. 
After training both networks, \texttt{PDE-READ} uses a modified version of the \emph{Recursive Feature Elimination} algorithm \cite{guyon2002gene} to extract $c_1, \ldots, c_K$ from $N$.
\cite{atkinson2019data} uses a Gaussian process to approximate the system response function and a genetic algorithm to identify the hidden PDE. 
Finally, \cite{bonneville2021bayesian} uses a Bayesian Neural Network to learn an approximation to the system response function and a sparse regression algorithm to recover the hidden PDE.

\section{Methodology}
\label{Sec:Method}

In this section, we describe the \texttt{weak-PDE-LEARN} algorithm in detail.
\texttt{weak-PDE-LEARN} relies on the weak form of the hidden PDE, equation \eqref{Eq:PDE} \cite{reddy2013introductory}. 
In subsection \ref{SubSec:Method:Weight} we show that specially selected \emph{weight functions} transform the weak form of the hidden PDE into a system of linear equations for the unknown coefficients, $c_1, \ldots, c_M$.
Following this motivation, we detail the \texttt{weak-PDE-LEARN} algorithm in subsection \ref{SubSec:Method:weak_PDE_LEARN}.

\subsection{Weight Functions and the Weak Form}
\label{SubSec:Method:Weight}

In this subsection, we will use a collection of infinitely differentiable, compactly supported weight (test) functions to derive a system of linear equations for $c_1, \ldots, c_M$.
The coefficients in this system of equations depend on the system response function, $u$, the weight functions, and their derivatives.
Crucially, by taking this approach, we do not need to evaluate $u$'s classical partial derivatives; thereby avoiding the pitfalls in computing derivatives from noisy observational data.

\subsubsection{The Weak Form of the Hidden PDE}
\label{SubSubSec:Method:Weak_Form}

Let $w_1, \ldots, w_K \in C_c^{\infty}\left([0, T] \times \Omega\right)$ be infinitely differentiable, compactly supported functions on the problem domain, $[0, T] \times \Omega$. 
We refer to these as the \emph{weight functions}.
Let $k \in \{ 1, 2, \ldots, K \}$. 
We will assume that the support of $w_k$ lies in the interior of $[0, T] \times \Omega$. 
In other words, $w_k|_{\partial \left( [0, T] \times \Omega \right)} = 0$.

\bigskip

Multiplying the hidden PDE, equation \eqref{Eq:PDE}, by $w_k$ yields

\begin{equation} w_k(t, X) D^{\alpha(0)} F_0\left(u\left(t, X\right) \right)= \sum_{m = 1}^{M} c_m w_k(t, X) D^{\alpha(m)} F_m\left(u\left(t, X\right) \right). \label{Eq:PDE:With_Weights} \end{equation}

Integrating over the problem domain, $[0, T] \times \Omega$ gives 

\begin{equation} \int_{[0, T] \times \Omega} w_k(t, X) D^{\alpha(0)} F_0\left(u\left(t, X\right) \right)\ dt\ d\Omega = \sum_{m = 1}^{M} c_m \int_{[0, T] \times \Omega} w_k(t, X) D^{\alpha(m)} F_m\big(u(t, X) \big)\ dt\ d\Omega. \label{Eq:PDE:Integrated} \end{equation}

Let's focus on one of the terms in the summation on the right:

$$\int_{[0, T] \times \Omega} w_k(t, X) D^{\alpha(m)} F_m\big(u(t, X) \big)\ dt\ d\Omega.$$

By assumption, $w_k$ is infinitely differentiable and has $w_k|_{\partial \left( [0, T] \times \Omega \right)} = 0$.
Therefore, by repeatedly applying Green's lemma \cite{langtangen2003computational} we must have

\begin{multline} 
\int_{[0, T] \times \Omega} w_k(t, X) D^{\alpha(m)} F_m\big(u(t, X) \big)\ dt\ d\Omega = \\ 
\left(-1 \right)^{|\alpha(m)|} \int_{[0, T] \times \Omega} \left( D^{\alpha(m)} w_k(t, X) \right) F_m\big(u(t, X) \big)\ dt\ d\Omega, \label{Eq:Integration_By_Parts}
\end{multline} 

where $|\alpha(m)|$ denotes the sum of the components in the multi-index $\alpha(m)$. 
Doing this for each $m$ (and on the left hand side) and substituting the results back into equation \eqref{Eq:PDE:Integrated} gives

\begin{multline}
\left(-1\right)^{|\alpha(0)|} \int_{[0, T] \times \Omega} \left( D^{\alpha(0)} w_k(t, X) \right) F_0\big(u(t, X)\big) \ dt\ d\Omega = \\
\sum_{m = 1}^{M} c_m \left(-1 \right)^{|\alpha(m)|} \int_{[0, T] \times \Omega} \left( D^{\alpha(m)} w_k(t, X) \right) F_m\big(u(t, X) \big)\ dt\ d\Omega. \label{Eq:PDE:Weak} 
\end{multline}

Doing this for each $k \in \{ 1, 2, \ldots, K \}$ gives a system of linear equations for $c_1, \ldots, c_M$. 
To make this system more concise, let $b(u) \in \mathbb{R}^K$ and $A(u) \in \mathbb{R}^{K \times M}$ be defined as follows

\begin{align} 
b(u)_k &= \left(-1\right)^{|\alpha(0)|} \int_{[0, T] \times \Omega} \left( D^{\alpha(0)} w_k(t, X) \right) F_0\big( u(t, X) \big)\ dt\ d\Omega \label{Eq:PDE:b(u)} \\
A(u)_{k,m} &= \left(-1 \right)^{|\alpha(m)|} \int_{[0, T] \times \Omega} \left( D^{\alpha(m)} w_k(t, X) \right) F_m\big(u(t, X) \big)\ dt\ d\Omega \label{Eq:PDE:A(u)}.
\end{align}

Then, the system of linear equations can be expressed as 

\begin{equation} A(u)c = b(u) \label{Eq:PDE:Ac_b} \end{equation}

where 

$$c = c_1 e_1 + \cdots + c_M e_M.$$ 

Here, $e_k$ is the $k$th standard basis vector in $\mathbb{R}^M$. 
Thus, $e_k$ is the $M$-component vector whose $k$th component is $1$ and whose other components are $0$.

\subsubsection{Product-of-Bump Weight Functions}
\label{SubSubSec:Method:Weight_Functions}

The results above hold for any collection $w_1, \ldots, w_K \in C^{\infty}_{c}([0, T] \times \Omega)$. 
However, in this paper, we chose our weight functions to be variations of the classic \emph{bump} function found throughout PDE theory \cite{evans2022partial}.
If the problem domain is a subset of $\mathbb{R} \times \mathbb{R}^{d}$, then each of our weight functions is a product of $d + 1$ bump functions.
More specifically, our weight functions have the general form

\begin{equation} w(t, X) = 
\begin{cases} \exp\left( \frac{\beta r^2}{(t - t_0)^2 - r^2} + \beta \right) \prod_{i = 1}^{d} \exp\left(\frac{\beta r^2}{([X]_i - [X_0]_{i})^2 - r^2} + \beta\right) & \text{if } (t, X) \in B^{\infty}_{r}\left(t_0, X_0\right) \\
0 & \text{otherwise} \end{cases}
\label{Eq:Weight_Function}
\end{equation}

where $B^{\infty}_{r}(t_0, X_0)$ is the $\infty$-norm ball of radius $r$ and center $(t_0, X_0)$. 
More formally, 

$$B^{\infty}_{r}\left( t_0, X_0 \right) = \left\{ (t, X) \in \mathbb{R}^{d + 1} : |t - t_0| < r \text{ and } \big| [X]_i - [X_0]_i \big| < r \text{ for } i = 1, 2, \ldots, d \right\}.$$

Here, $\beta > 0$ is a user-selected constant that controls how quickly the weight function decays to zero.

\bigskip

To build the weight functions, we first sample $K$ points from the interior of $[0, T] \times \Omega$. 
Let $(t_{0, 1}, X_{0, 1}), \ldots, (t_{0, K}, X_{0, K})$ denote the resulting collection of points.
We then find radii $r_0, \ldots, r_K$ such that for each $k$, $B^{\infty}_{r_k}(t_{0,k}, X_{0,k}) \subseteq [0, T] \times \Omega$.
We then define the weight functions using the resulting centers and radii.
One particular advantage of this approach is that it allows us to streamline the computations required to evaluate the partial derivatives of the weight function.
We discuss this in detail in appendix \ref{Appendix:Master_Weight_Function}. 
In our experiments, we arbitrarily select $\beta = 5$.

\subsubsection{Significance of the Weak Form Approach}
\label{SubSubSec:Method:Significance}

Many existing PDE discovery algorithms (see section \ref{Sec:Related_Work}) require evaluating $u$ and its partial derivatives.
Often, however, we only have access to noisy measurements of the system response function.  
In principle, we can then use finite differences (or other numerical differentiation approaches) to approximate their derivatives.
However, differentiation tends to amplify noise \cite{liu2003computational}.
Thus, learning an approximation to $u$ and its partial derivatives can be very challenging if we only have noisy measurements of $u$.

\bigskip

Equations \eqref{Eq:PDE:b(u)} and \eqref{Eq:PDE:A(u)} are significant because they do not involve the partial derivatives of the system response function, $u$. 
They only involve the partial derivatives of the weight functions.
However, since we select the weight functions, we can compute these partial derivatives exactly.
Therefore, if we can learn an approximation of the system response function, we can use that approximation to evaluate the linear system in equation \eqref{Eq:PDE:Ac_b}.
Not needing to approximate partial derivatives gives weak-form approaches a key advantage over many of those discussed in section \ref{Sec:Related_Work}. 
We discuss this advantage in greater detail in section \ref{Sec:Discussion}. 

\bigskip

If $u$ satisfies equation \eqref{Eq:PDE}, then for $K \geq M$ (more rows than columns), there should be a unique solution to equation \eqref{Eq:PDE:Ac_b}.
This solution gives the coefficients in the hidden PDE, equation \eqref{Eq:PDE}.
In our approach, however, we do not know a solution to the hidden PDE; we only have access to limited, noisy measurements that presumably emanate from one.
As such, we will generally work with a function $U : [0, T] \times \Omega \to \mathbb{R}$ that approximately satisfies the hidden PDE in the sense that the left and right-hand sides of equation \eqref{Eq:PDE:With_Weights} are approximately equal for $t, X \in [0, T] \times \Omega$ and $k = 1, 2, \ldots, K$.
As a result, we can only expect $A(U)c \approx b(U)$.
For this reason, we seek a sparse least-squares solution to equation \eqref{Eq:PDE:Ac_b}.

\bigskip

Finally, notice that the components of $A(u)$ and $b(u)$ are all expressed in terms of integrals. 
In subsection \eqref{SubSec:Method:Weight}, we outline how to use neural networks to learn an approximation, $U : [0, T] \times \Omega \to \mathbb{R}$, of the system response function.
We then use $U$ to numerically approximate (using the Trapezoidal rule) the integrals necessary to compute the components of $A(U)$ and $b(U)$. 
Since we train $U$ on noisy measurements of the system response function, we expect $U$ to be a noisy approximation of the true solution to the hidden PDE, $u$. 
However, since we assume the noise has a mean of zero, we can expect the integrals to average out this noise.
In other words, the nature of the equations used to evaluate the components of $A(U)$ and $b(U)$ should make the solution to the corresponding linear system robust to noise.

\bigskip 

Table \ref{Table:Method:Weight:Notation} lists the notation and terminology we introduced in the foregoing subsection.

\begin{table}[hbt]
    \centering 
    \rowcolors{2}{cyan!10}{white}
    
    \begin{tabulary}{0.9\linewidth}{p{3cm}L}
        \toprule[0.3ex]
        \textbf{Notation} & \textbf{Meaning} \\
        \midrule[0.1ex]
        $|\alpha(m)|$ & The sum of the indices in a multi-index, $\alpha(m)$. \\
        \addlinespace[0.4em]
        $K$ & The number of weight functions. \\
        \addlinespace[0.4em]
        $w_k$ & The $k$th weight function. This is an infinitely differentiable, compactly supported function, defined on $[0, T] \times \Omega$, which satisfies $w_k|_{\partial \Omega} = 0$. In \texttt{weak-PDE-LEARN}, these are defined by equation \eqref{Eq:Weight_Function}. \\
        \addlinespace[0.4em]
        $\beta$ & A constant used in the definition of our weight functions. See equation \eqref{Eq:Weight_Function}. In this paper, $\beta = 5$. \\
        \addlinespace[0.4em]
        $r, (X_0, t_0)$ & Constants representing the radius and center (in both components of the problem domain) of a weight function, respectively. See equation \eqref{Eq:Weight_Function}. \\
        \addlinespace[0.4em]
        $A(u)$ & A $K \times M$ matrix whose $k,m$ entry is $\left(-1 \right)^{|\alpha(m)|} \int_{[0, T] \times \Omega} \left( D^{\alpha(m)} w_k(t, X) \right) F_m\big(u(t, X) \big)\ dt\ d\Omega$. \\
        \addlinespace[0.4em]
        $b(u)$ & A $K$ element vector whose $k$th component is $(-1)^{|\alpha(0)|} \int_{[0, T] \times \Omega} \left( D^{\alpha(0)} w_k(t, X) \right) F_0\big(u(t, X))\ dt\ d\Omega$. \\
        $c$ & An element of $\mathbb{R}^M$ whose $m$th component is the coefficient $c_m$ from equation \eqref{Eq:PDE}. \\
        \bottomrule[0.3ex]
    \end{tabulary}
    
    \caption{The notation and terminology introduced in section \ref{SubSec:Method:Weight}} 
    \label{Table:Method:Weight:Notation}
\end{table}

\subsection{weak-PDE-LEARN}
\label{SubSec:Method:weak_PDE_LEARN}

\texttt{weak-PDE-LEARN} uses the noisy data set, $\{ \tilde{u}(t_i, X_i) \}_{i = 1}^{N_{Data}}$, to simultaneously learn an approximation of the system response function and the coefficients $c_1, \ldots, c_M$ in the hidden PDE.
More specifically, we train a Rational Neural Network (RatNN) \cite{boulle2020rational}, $U : [0, T] \times \Omega \to \mathbb{R}$, to match the noisy data set, $\{ \tilde{u}(t_i, X_i) \}_{i = 1}^{N_{Data}}$ and learn an approximation to the system response function.
A rational neural network is a standard, fully connected network whose activation functions are trainable type $3, 2$ rational functions.
Thus, the activation function looks like $p(x)/q(x)$, where $p$ is a third-order polynomial and $q$ is a second-order one.
The coefficients in each polynomial are trainable parameters that our network learns together with the weights and biases of each layer. 
Each layer gets a rational activation function with its own set of trainable coefficients (we apply the activation function component-wise to the output of that layer).

\bigskip 

While $U$ trains, we simultaneously train a vector $\xi \in \mathbb{R}^M$ to approximate the coefficient vector $c = c_1 e_1 + \cdots + c_M e_M$. 
\bigskip

\texttt{weak-PDE-LEARN} learns $U$ and $\xi$ by minimizing the following loss function:

\begin{equation} \begin{aligned} \text{Loss}\big( U, \xi \big) = \lambda_{\text{Data}} \text{Loss}_{\text{Data}}\big( U \big) + \lambda_{\text{Weak}} \text{Loss}_{\text{Weak}}\big(U, \xi\big) + \lambda_{L^p} \text{Loss}_{L^p}(\xi). \end{aligned} \label{Eq:Loss} \end{equation}

Here, $\text{Loss}_{\text{Data}}$, $\text{Loss}_{\text{Weak}}$, and $\text{Loss}_{L^p}$ are the \emph{data, weak form}, and \emph{$L^p$} losses, respectively.
Further, $\lambda_{\text{Data}}$, $\lambda_{\text{Weak}}$, and $\lambda_{L^p}$ are positive, user-defined constants that control the relative weight of the three parts of the loss function.
In all of our experiments (see section \ref{Sec:Experiments}), we use $\lambda_{\text{Data}} = \lambda_{\text{Weak}} = 1.0$.
We do not claim these values are optimal and recommend treating each weight as a tunable hyper-parameter when using \texttt{weak-PDE-LEARN} in practice.
We now define and discuss each of the three loss functions.

\bigskip

\textbf{Data Loss:} $\text{Loss}_{\text{Data}}$ is the mean-square error between the network's predictions at the data points, $\{ (t_i, X_i) \}_{i = 1}^{N_{Data}}$, and the noisy data set, $\{ \tilde{u}(t_i, X_i) \}_{i = 1}^{N_{Data}}$:

\begin{equation} \text{Loss}_{\text{Data}}\big(U) = \left( \frac{1}{N_{Data}} \right) \sum_{i = 1}^{N_{Data}} \Big| U\left(t_{i}, X_{i} \right)  - \tilde{u}\left(t_{i}, X_{i} \right) \Big|^2 \label{Eq:Loss:Data} \end{equation}

Thus, $\text{Loss}_{\text{Data}}$ quantifies how well $U$ matches the noisy measurements of the system response.

\bigskip

\textbf{Weak Form Loss:} $\text{Loss}_{\text{Weak}}$ quantifies how well $U$ satisfies the hidden PDE when the components of $\xi$ replace those of $c$. 
More specifically, using the notation of subsection \eqref{SubSec:Method:Weight},

\begin{equation}
\begin{aligned} \text{Loss}_{\text{Weak}}(U, \xi) &= \sum_{k = 1}^{K} \Bigg| \left(-1\right)^{|\alpha(0)|} \int_{[0, T] \times \Omega} \left( D^{\alpha(0)} w_k \right) F_0(U) \ dt\ d\Omega\ -\\
&\qquad\qquad \sum_{m = 1}^{M} \xi_m (-1)^{|\alpha(m)|} \int_{[0, T] \times \Omega} \left( D^{\alpha(m)} w_k \right) F_m\big(U\big)\ dt\ d\Omega \Bigg|^2 \\
&= \sum_{k = 1}^{K} \Big(b(U)_k - A(U)[k, :]\xi \Big)^2 \\
&= \Big\| b(U) - A(U)\xi \Big\|_2^2.
\end{aligned}
\label{Eq:Loss:Weak}
\end{equation}

We use the multi-dimensional trapezoidal rule to approximate all of the integrals in equation \eqref{Eq:Loss:Weak}.
Importantly, the weak form loss couples the training of $\xi$ and $U$.
It is merely the least-squares loss of the linear system from equation \eqref{Eq:PDE:Ac_b}.
Thus, including it in the loss encourages $\xi$ and $U$ to satisfy the linear system $A(U)\xi = b(U)$ in the least squares sense.
Since this linear system is derived from the hidden PDE, this part of the loss function trains $U$ and $\xi$ to satisfy the hidden PDE. 

\bigskip

\textbf{L\textsuperscript{p} Loss:} The third and final component of the loss function is defined by

\begin{equation} \text{Loss}_{L^p}(\xi) = \sum_{m = 1}^{M} \eta_m |\xi_m|^2. \label{Eq:Loss:Lp} \end{equation}

Here $p > 0$ is a user-specified hyperparameter and  

\begin{equation} \eta_m = \frac{1}{\max\left(|\xi_m|^{2 - p}, \delta \right)}, \label{Eq:Loss:Lp:eta} \end{equation}

where $\delta > 0$ ensures we avoid division by zero. We use $\delta = 1\text{e-}7$.
As a consequence of this definition, we can conclude that if the components of $\xi$ are sufficiently large, then 

\begin{equation}\text{Loss}_{L^p}(\xi) = \sum_{m = 1}^{M} |\xi_m|^p. \label{Eq:Loss:Lp:2}\end{equation}

Critically, however, $\eta_m$ is treated as a constant during training (not a function of $\xi_m$).
We re-calculate $\eta_m$ at the start of each training epoch but treat it as a constant throughout the step.
This difference is crucial; equation \eqref{Eq:Loss:Lp} is strictly convex while equation \eqref{Eq:Loss:Lp:2} is not and is, therefore, more challenging to minimize via gradient descent. 
This trick, inspired by \emph{iteratively re-weighted least squares}, \cite{chartrand2008iteratively}, effectively allows us to penalize \eqref{Eq:Loss:Lp:2} for $p$ smaller than $1$.
By choosing a value of $p$ close to zero and exploiting the fact that

$$\lim_{p \to 0^+} |x|^p = 1_{\{x\ \neq\ 0\}},$$ 

we have 

$$\text{Loss}_{L^p}(\xi) \approx \| \xi \|_0 = \text{ Number of non-zero terms in } \xi.$$

In other words, it allows us to enforce sparsity in $\xi$. 
For our experiments, we arbitrarily chose $p = 0.1$.

\subsubsection{Adaptive Selection of the Weight Functions}
\label{SubSubSec:Method:Adaptive_Weights}

Here, we describe an adaptive procedure that \texttt{weak-PDE-LEARN} uses to select the weight functions.
We based this approach on the algorithm \texttt{PDE-LEARN} uses to adaptively select its \emph{targeted collocation points} \cite{stephany2022LEARN}.

\bigskip 

During training, \texttt{weak-PDE-LEARN} uses two kinds of weight functions: \emph{random weight functions} and \emph{targeted weight functions}.
\texttt{weak-PDE-LEARN} samples new random weight functions periodically.
In our numerical experiments, we did this resampling every $20$ epochs.

\bigskip 

To sample a new set of random weight functions, we first sample $N_{\text{Random}}$ points from the problem domain, $[0, T] \times \Omega$, by repeatedly sampling a uniform distribution whose support is the problem domain.
Here, $N_{\text{Random}}$ is a user-selected hyper-parameter that sets the number of random weight functions \footnote{When we re-sample the random weight functions, we discard the old ones and replace them with the ones we just created.}.
In our experiments, we use $N_{\text{Random}} = 200$.
The resulting collection of $N_{\text{Random}}$ points acts as the centers for the new random weight functions.
For each sampled point, $(t, X)$, we find a radius $r > 0$ such that $B^{\infty}_r(t, X) \subseteq [0, T] \times \Omega$. 
These numbers become the radii for the new random weight functions.

\bigskip 

To compute the weak form loss, \texttt{weak-PDE-LEARN} must evaluate the partial derivatives of the weight functions at the quadrature points used to approximate the integrals in equation \eqref{Eq:Loss:Weak}.
Thus, after we find the centers and radii of the new random weight functions, the next step is to compute their partial derivatives.
In principle, we could directly compute these derivatives by directly differentiating the weight functions (see equation \eqref{Eq:Weight_Function}).
Such an approach would involve unnecessary computation, however.
A better solution is to define a \emph{master weight function} and then define all other weight functions relative to it.
This approach works because of a key mathematical property of the weight functions:
given any pair of weight functions --- $w$ and $\tilde{w}$ --- we can express $\tilde{w}$ as the composition of $w$ with an affine map.
This result means we can write the partial derivatives of each weight function in terms of those of the master weight function.
Reusing the computed partial derivatives allows us to considerably reduce the computational effort required to define a new weight function.
We detail this approach in Appendix \ref{Appendix:Master_Weight_Function}.

\bigskip 

The set of targeted weight functions is initially empty, but \texttt{weak-PDE-LEARN} updates it at the end of each epoch.
During each epoch, we evaluate the weak form loss, equation \eqref{Eq:Loss:Weak}, at the current set of random and targeted weight functions.
Thus, each row of $A(U)\xi - b(U)$ in equation \eqref{Eq:Loss:Weak} corresponds to either a targeted or random weight function.
In principle, if $U$ is a good approximation of the system response function, $u$, and $\xi$ is a good approximation of the true coefficient vector, $c$, then each element of $A(U)\xi - b(U)$ will be small. 
If some of these components have a large magnitude, this suggests that something can be improved (either $U$ is a poor approximation of $u$, or $\xi$ is a poor approximation of $c$, or both).
We use this information to focus \texttt{weak-PDE-LEARN}'s training.
In particular, at the end of each epoch, we use the following procedure to update the set of targeted weight functions:

\begin{enumerate}
\item \texttt{weak-PDE-LEARN} records the absolute value of each component of $A(U)\xi - b(U)$. 
\item \texttt{weak-PDE-LEARN} then computes the mean and standard deviation of this set.
\item \texttt{weak-PDE-LEARN} then determines which components of $A(U)\xi - b(U)$ have an absolute value more than two standard deviations greater than the mean. 
These corresponding weight functions (remember that each row corresponds to a weight function) become the set of targeted weight functions in the next epoch.
\end{enumerate}

This approach helps to accelerate the training of $U$ and $\xi$ by holding onto the weight functions that engendered a large residual (a component of $A(U)\xi - b(U)$).
These weight functions are, in essence, the ones that reveal the biggest opportunities for improvement in the current $U$ and $\xi$.
Keeping them for future epochs helps the optimizer train $U$ and $\xi$ to become better approximations of $u$ and $c$, respectively.

\subsubsection{Minimizing the Loss Function}
\label{SubSubSec:Method:Minimizing_Loss}

We minimize the loss, equation \eqref{Eq:Loss}, using a combination of the \texttt{Adam} \cite{kingma2014adam} and \texttt{LBFGS} \cite{liu1989limited} optimizers. 
We do this using the same three-step training procedure in \cite{stephany2022LEARN}.

\bigskip

During the \emph{burn-in} step, we train $\xi$ and $U$ with $\lambda_{L^p} = 0$. 
During this period, $U$ uses the noisy data set to approximate the system response function while satisfying a PDE of the general form of the hidden PDE, equation \eqref{Eq:PDE}.
The weak form loss acts as a powerful regularizer during this step, preventing $U$ from over-fitting the noisy data set.

\bigskip 

During the \emph{sparsification} step, we eliminate all components of $\xi$ which were smaller than $\sqrt{\delta}$ (from equation \eqref{Eq:Loss:Lp:eta}) at the end of the burn-in step\footnote{\texttt{weak-PDE-LEARN} can not resolve components with smaller values because of floating point precision see \cite{stephany2022LEARN}.}.
We then resume training with $\lambda_{L^p}$ set to a small, but non-zero, value.
During this step, $\xi$ learns a sparse approximation to $c$.
This step is when our method identifies which terms are vital to retain (non-zero coefficient) in the hidden PDE.
At the end of the sparsification step, the coefficients tend to be too small (since the $L^p$ loss encourages the components of $\xi$ to approach zero).

\bigskip

The third and final step is the \emph{fine-tuning} step.
We begin by eliminating all components of $\xi$ which were smaller than $\sqrt{\delta}$ and then resume training with $\lambda_{L^p} = 0$. 
During this step, the $L^p$ loss increases (as the optimizer is no longer trying to minimize it).
Training continues until the $L^p$ loss stops increasing.
We then report the PDE encoded in $\xi$. 

\bigskip

In our experiments, the first two steps always use the \texttt{Adam} optimizer, while the final step uses the \texttt{LBFGS} optimizer.
We let $N_{Burn}$, $N_{Sparse}$, and $N_{Tune}$ denote the number of burn-in, sparsification, and fine-tuning epochs, respectively.

\bigskip

Table (\ref{Table:Method:weak_PDE_LEARN:Notation}) lists the notation we introduced in the foregoing section.

\begin{table}[hbt]
    \centering 
    \rowcolors{2}{white}{cyan!10}
    
    \begin{tabulary}{0.9\linewidth}{p{3.5cm}L}
        \toprule[0.3ex]
        \textbf{Notation} & \textbf{Meaning} \\
        \midrule[0.1ex]
        $U$ & A RatNN that learns an approximation of the system response function $u$. \\
        \addlinespace[0.4em]
        $\xi$ & An trainable, $M$ element vector that learns to approximate $c$. \\
        \addlinespace[0.4em]
        $\text{Loss}_{\text{Data}}, Loss_{\text{Weak}}$, $\text{Loss}_{L^p}$ & The data, weak form, and $L^p$ losses, respectively. See equations \eqref{Eq:Loss}, \eqref{Eq:Loss:Data}, \eqref{Eq:Loss:Lp}, and \eqref{Eq:Loss:Weak}, respectively. \\
        \addlinespace[0.4em]
        $\lambda_{\text{Data}}$, $\lambda_{\text{Weak}}$, $\lambda_{L^p}$ & The weights of the Data, Weak, and $L^p$ parts of the loss function. See equation \eqref{Eq:Loss}. \\
        \addlinespace[0.4em]
        $N_{\text{Random}}$ & The number of random weight functions. We periodically re-sample these using the procedure outlined in subsubsection \ref{SubSubSec:Method:Adaptive_Weights}. In our experiments, we use $N_{\text{Random}} = 200$. \\
        \addlinespace[0.4em]
        $N_{Burn}$, $N_{sparse}$, $N_{Tune}$ & The number of of burn-in, sparsification, and fine-tuning epochs, respectively. \\
        \addlinespace[0.4em]
        $p$ & A user-specified hyperparameter that specifies the ``$p$'' in the $L^p$ loss. See equation \eqref{Eq:Loss:Lp}. In our experiments, we use $p = 0.1$. \\
        \bottomrule[0.3ex]
    \end{tabulary}
    
    \caption{The notation and terminology introduced in section \ref{SubSec:Method:weak_PDE_LEARN}} 
    \label{Table:Method:weak_PDE_LEARN:Notation}
\end{table}
\section{Experiments}
\label{Sec:Experiments}

In this section, we test \texttt{weak-PDE-LEARN} on several benchmark PDEs. 
In particular, we demonstrate that \texttt{weak-PDE-LEARN} can identify the Burgers, KdV, and KS equations from noisy, limited measurements of solutions to these equation.
To run these experiments, we implemented \texttt{weak-PDE-LEARN} as an open-source Python library.
Our implementation is available at \url{https://github.com/punkduckable/Weak_PDE_LEARN}.
It uses \texttt{Pytorch} \cite{paszke2019pytorch} to define and train $U$ and $\xi$ \cite{paszke2017automatic}.

\bigskip 

Notably, our implementation is slightly more general than the algorithm in section \ref{Sec:Method}.
In particular, it can train using several data sets simultaneously.
In this case, we assume $i$th data set represents noisy measurements of some function $u_i$.
We assume that $u_i$ satisfies the \emph{same} PDE, \eqref{Eq:PDE}, with the \emph{same} coefficients for each $i$.
If the user supplies multiple data sets, our implementation defines and trains a network for each data set but uses a common $\xi$ to encode the common hidden PDE.
Each data set also gets its own set of weight functions (defined on the associated problem domains).
During training, we compute the data and weak form losses for each network.
To find the targeted weight functions, we apply the adaptive weight function scheme described in subsection \ref{SubSec:Method:weak_PDE_LEARN} independently to each data set.
We then replace $\text{Loss}_{\text{Data}}$ and $\text{Loss}_{\text{weak}}$ in section \ref{SubSec:Method:weak_PDE_LEARN} with the sum of the data and weak form losses computed for each network, respectively.

\bigskip 

All our experiments deal with PDEs of two variables (one temporal, one spatial).
We chose this because we can plot the solutions, which helps us analyze \texttt{weak-PDE-LEARN}’s performance in these experiments.
Since the purpose of this paper is to introduce the \texttt{weak-PDE-LEARN} algorithm, we felt that restricting our attention to PDEs with two variables was the correct pedagogical choice.
Notwithstanding this focus on one spatial and one temporal variable, our implementation can identify PDEs with up to four variables (one for time and up to three for space).
Further, we wrote our code such that extending it to work with higher dimensional data sets is straightforward.

\bigskip 

\textbf{Generating the Data Sets:}
We used numerical simulations to generate all of the data in our experiments.
For each PDE, we use \texttt{Chebfun}'s \cite{driscoll2014chebfun} \texttt{spin} class to find an approximate, numerical solution to that PDE on a rectangular domain.
After solving, we evaluate the approximate solution on a rectangular grid of points.
The resulting set of data is our noise-free data set.
We then use the following procedure to make a noisy, limited (sub-sampled) data set with $N_{Data}$ points and a noise level of $q \geq 0$ (See section \ref{SubSec:System_Response}):

\begin{enumerate}
    \item First, calculate the standard deviation, $\sigma_{nf}$, of the noise-free data set.
    \item Randomly select a subset of size $N_{Data}$ from the noise-free data set by drawing $N_{Data}$ samples from the noise-free data set without replacement. 
    This process yields the \emph{limited data set}.
    \item For each point in the limited data set, sample a Gaussian Distribution with mean $0$ and standard deviation $q\, \sigma_{nf}$. 
    Add the $i$th resulting value to the $i$th element of the limited data set.
    The resulting set is our noisy, limited data set.
\end{enumerate}

\bigskip

\textbf{Fixed Hyperparameter values:} 
In all our experiments, we re-sampled the random-weight functions every $20$ epochs.
We also use $200$ random weight functions ($N_{\text{Random}} = 200$).
We also somewhat arbitrarily chose to use $\beta = 5$ to define our weight functions (see equation \eqref{Eq:Weight_Function}).
Further, in all experiments, we use $p = 0.1$, $\lambda_{\text{Data}} = 1.0$, and $\lambda_{\text{Weak}} = 1.0$ in the loss function (see equations \eqref{Eq:Loss} and \eqref{Eq:Loss:Lp}).
$\lambda_{\text{L}^p}$ changes according to the procedure outlined in sub-section \ref{SubSubSec:Method:Minimizing_Loss}

\bigskip 

Notably, we did not attempt to optimize the hyperparameters in the previous paragraph; they worked well for our tests, but we do not claim they are  optimal.
In practice, it may be worth adjusting some of these values to see what works best for a particular problem.

\bigskip 

We also used the same architecture for $U$ in all experiments.
Specifically, $U$ is a five-layer Rational Neural Network \cite{boulle2020rational} with $40$ neurons per layer.

\bigskip 

Finally, in all our experiments, the left-hand side term is $D_t U$, while the right-hand side terms are the following:

\begin{align*}
1,\ U,\ D_x &U,\ D_x^2 U,\ D_x^3 U,\ D_x^4 U, \\
U^2,\ D_x &U^2,\ D_x^2 U^2,\ D_x^3 U^2, \\
U^3,&\ D_x U^3,\ D_x^2 U^3 \\
\end{align*}

\begin{figure}[!hbt]
    \centering
    \includegraphics[width=.8\linewidth]{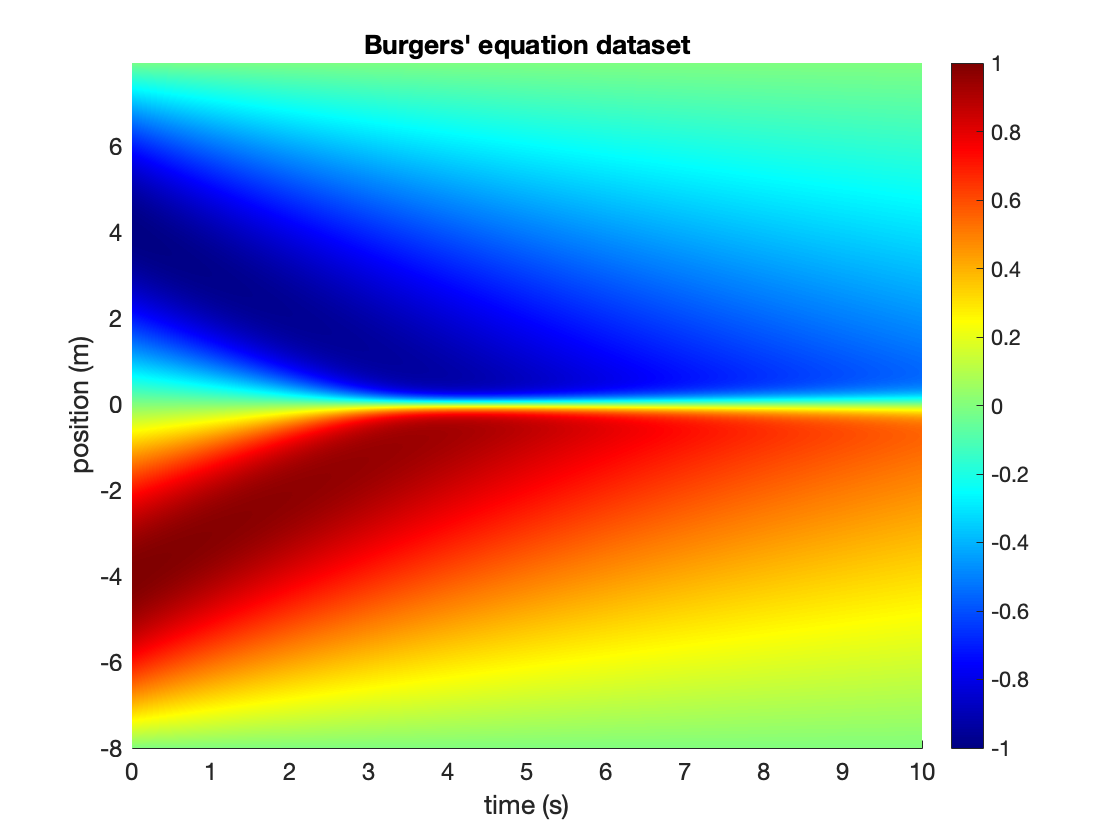}
    \caption{Noise-free Burgers' equation data set.}
    \label{Fig:Experiments:Burgers:Dataset}
\end{figure}

\subsection{Burgers Equation}
\label{SubSec:Experiments:Burgers}

Burgers' equation is a second-order, non-linear PDE. 
The equation first appeared in the context of Fluid Mechanics \cite{bateman1915some} but has since found application in Non-linear Acoustics, Gas Dynamics, and Traffic Flow \cite{basdevant1986spectral}.
Burgers equation is 

$$D_t u = \nu \left( D_x^2 u \right) + 0.5 \left(D_x \left( u^2 \right) \right),$$

where $\nu > 0$ is the {\it diffusion coefficient}.
We can interpret $u(t, x)$ as the velocity of a fluid flow at $(t, x) \in [0, T] \times \Omega$.

\bigskip 

To test \text{weak-PDE-LEARN} on Burgers' equation, we first use \texttt{Chebfun} to generate a collection of noisy, limited data sets.
For this experiment, $\nu = 0.1$ and $[0, T] \times \Omega = [0, 10] \times [-8, 8]$. 
We also use the following initial condition:

$$u(0, x) = -\sin\left( \pi \frac{x}{8} \right)$$

Our script \texttt{Burgers\_Sine.m} (in the \texttt{MATLAB} sub-directory of our repository) uses \texttt{Chebfun}'s \cite{driscoll2014chebfun} \texttt{spin} class to generate the noise-free data set for Burgers' equation.
Figure \ref{Fig:Experiments:Burgers:Dataset} depicts the noise-free data set.
We use the procedure outlined at the start of this section to generate several noisy and limited data sets.
We then test \texttt{weak-PDE-LEARN} on each data set.
Table \ref{Table:Experiments:Burgers} reports the results of these experiments.
Note that the $\lambda_{L^p}$ column specifies the value we used for this hyperparameter during the sparsification step.

\begin{table}[hbt]
    \centering 
    \rowcolors{2}{cyan!10}{white}
    \begin{threeparttable}
        \caption{Experimental results for Burgers' equation} 
        \label{Table:Experiments:Burgers}

        \begin{tabulary}{\linewidth}{p{1.0cm}p{0.9cm}p{1.0cm}p{1.2cm}p{1.2cm}p{1.0cm}L}
            \toprule[0.3ex]
            $N_{Data}$ & Noise & $N_{Burn}$ & $N_{Sparse}$ & $\lambda_{L^p}$ & $N_{Tune}$ & Identified PDE \\
            \midrule[0.1ex] 
            $4,000$ & $25\%$ & $2,000$ & $2,000$ & $0.00002$ & $0$\tnote{\S} & $D_t U =  0.1013(D_x^2 U) -  0.5065(D_x U^2)$ \\
            \addlinespace[0.4em]
            $4,000$ & $50\%$ & $2,000$ & $2,000$ & $0.00002$ & $0$\tnote{\dag} & $D_t U =  0.0940(D_x^2 U) -  0.4596(D_x U^2)$ \\
            \addlinespace[0.4em]
            $4,000$ & $75\%$ & $2,000$ & $2,000$ & $0.0001$ & $100$ & $D_t U =  0.0847(D_x^2 U) -  0.4523(D_x U^2)$ \\
            \addlinespace[0.4em]
            $4,000$ & $100\%$ & $2,000$ & $1,000$ & $0.0001$ & $0$\tnote{\dag} & $D_t U =  0.0660(D_x^2 U) -  0.4411(D_x U^2)$ \\
            \bottomrule[0.3ex]
        \end{tabulary}

        \begin{tablenotes}
            \item [\S] Running fine-tuning epochs on the $25\%$ experiment didn't significantly change the coefficients as they were already nearly perfect.
            Thus, we report $0$ fine-tuning epochs for this experiment.
            
            \item [\dag] In these experiments, the $L^p$ loss decreased immediately during the fine-tuning step as $U$ began to over fit the data set. 
            Thus, we report the result after the sparsification step.
        \end{tablenotes}
    \end{threeparttable}
\end{table}

\bigskip

From table \ref{Table:Experiments:Burgers}, we see that \texttt{weak-PDE-LEARN} successfully learns Burgers' equation in all four experiments.
It can identify Burgers equation even when we contaminate the sparse and noisy data set with $100$\% noise.
These results match the performance of \texttt{PDE-LEARN} \cite{stephany2022LEARN}.
Further, the accuracy of the learned coefficients decreases as the noise level increases.

\bigskip 

\begin{figure}
\centering
\begin{subfigure}{.49\linewidth}
  \centering
  \includegraphics[width=\linewidth]{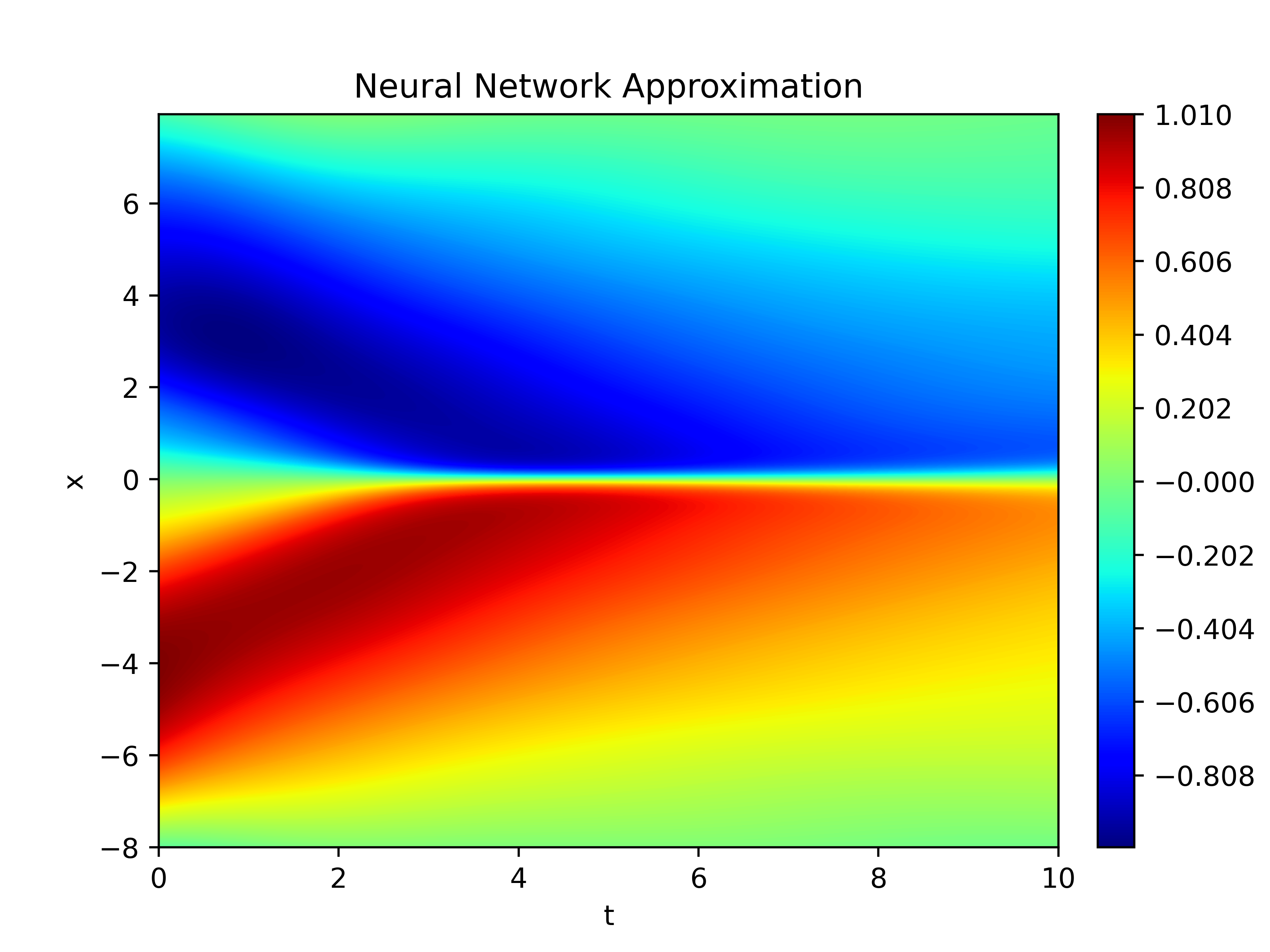}
  \caption{$U$ in the $25$\% noise experiment.}
  \label{SubFig:Experiments:Burgers:U25}
\end{subfigure}
\begin{subfigure}{.49\linewidth}
  \centering
  \includegraphics[width=\linewidth]{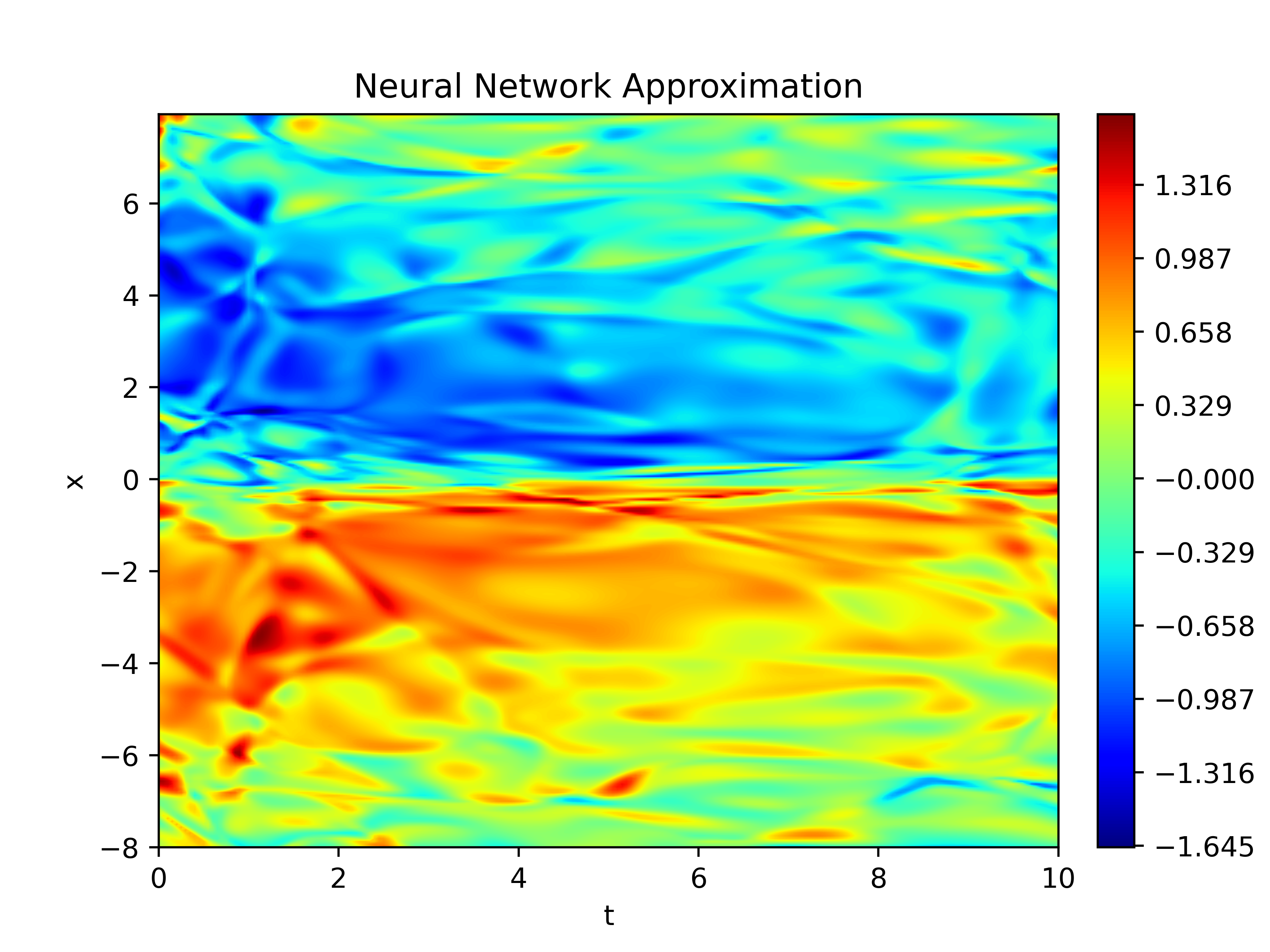}
  \caption{$U$ in the $75$\% noise experiment.}
  \label{SubFig:Experiments:Burgers:U75}
\end{subfigure}
\caption{Comparison of $U$ in the $25$\% and $75$\% noise Burgers' equation experiments}
\label{Fig:Experiments:Burgers:U}
\end{figure}

Figure \ref{Fig:Experiments:Burgers:U} depicts $U$ from the $25$\% and $75$\% noise experiments.
Interestingly, while $U$ from the $25$\% noise experiment closely resembles the noise-free data set, Figure \ref{Fig:Experiments:Burgers:Dataset}, $U$ in the $75$\% noise experiment does not.
The latter has some common characteristics with the noise-free data set, but also contains sharp lines and other irregular features that are not present in the true PDE solution.
Nonetheless, \texttt{weak-PDE-LEARN} can still confidentially identify Burgers' equation from Figure \ref{SubFig:Experiments:Burgers:U75}.
Remember that \texttt{weak-PDE-LEARN} identifies PDEs using the weak form of the hidden PDE, equation \ref{Eq:PDE:Integrated}.
While Figure \ref{SubFig:Experiments:Burgers:U75} looks quite different from the noise-free Burgers' equation data set, we can assume that the learned function behaves similar to the true solution when integrated against our weight functions.

\begin{figure}[!hbt]
    \centering
    \includegraphics[width=.8\linewidth]{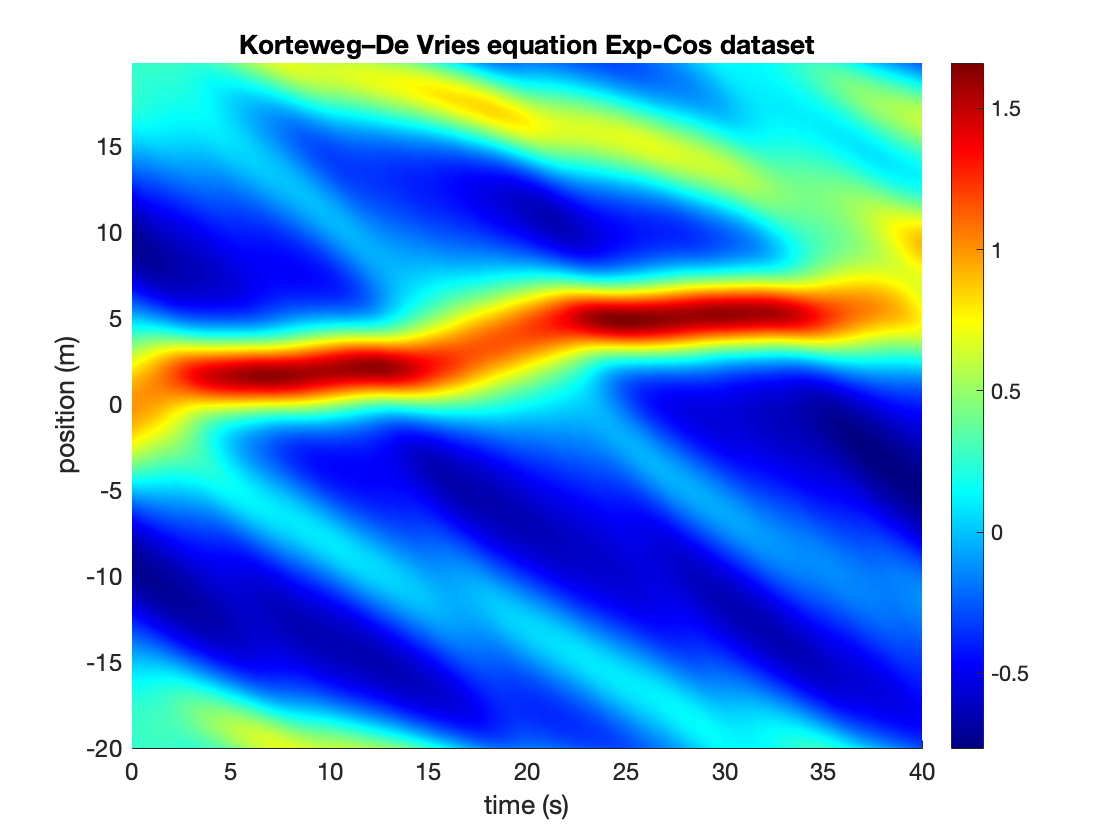}
    \caption{Noise-free KdV equation data set.}
    \label{Fig:Experiments:KdV:Dataset}
\end{figure}

\subsection{Korteweg–De Vries Equation}
\label{SubSec:Experiments:KdV}

The Korteweg-De Vries (KdV) equation is a third-order, non-linear equation.
\cite{korteweg1895xli} first proposed the equation to describe the motion of one-dimensional, shallow water waves. 
The KdV equation is

$$D_t u = - D_x^3 u - 0.5  \left( D_x \left( u^2 \right) \right).$$

Here, we can interpret $u(t, x)$ as the wave height at the point $(t, x) \in [0, T] \times \Omega$.

\bigskip

To test \texttt{weak-PDE-LEARN} on the KdV equation, we use \texttt{Chebfun} to generate noisy, limited data sets. 
In our experiments, $[0, T] \times \Omega = [0, 40] \times [-20, 20]$. 
We also use the following initial condition:

$$u(0, x) = \exp\left(-\pi\left(\frac{x}{30} \right)^2\right) \cos\left( \pi\frac{x}{10}\right) $$

The script \texttt{KdV\_Exp\_Cos.m} in the \texttt{MATLAB} sub-directory of our repository uses \texttt{Chebfun}'s \texttt{spin} class to generate the noise-free data sets. 
Figure \ref{Fig:Experiments:KdV:Dataset} depicts the noise-free data set for the KdV equation. 
We then use the procedure outlined at the start of this section to generate the noisy and limited data sets.
We then test \texttt{weak-PDE-LEARN} on each data set.
Table \ref{Table:Experiments:KdV} reports the results of those experiments.

\begin{table}[hbt]
    \centering 
    \rowcolors{2}{white}{cyan!10}
    \begin{threeparttable}
        \caption{Experimental results for KdV equation} 
        \label{Table:Experiments:KdV}

        \begin{tabulary}{\linewidth}{p{1.0cm}p{0.9cm}p{1.0cm}p{1.2cm}p{1.0cm}p{1.0cm}L}
            \toprule[0.3ex]
            $N_{Data}$ & Noise & $N_{Burn}$ & $N_{Sparse}$ & $\lambda_{L^p}$ & $N_{Tune}$ & Identified PDE \\
            \midrule[0.1ex] 
            $4,000$ & $25\%$ & $2,000$ & $1,000$ & $0.005$ & $100$ & $D_t U =  -0.9981(D_x^3 U) -  0.5035(D_x U^2)$ \\
            \addlinespace[0.4em]
            $4,000$ & $50\%$ & $2,000$ & $1,000$ & $0.005$ & $100$ & $D_t U =  -  0.9738(D_x^3 U) -  0.4876(D_x U^2)$ \\
            \addlinespace[0.4em]
            $4,000$ & $75\%$ & $2,000$ & $1,000$ & $0.005$ & $30$ & $D_t U =  -  0.9513(D_x^3 U) -  0.4740(D_x U^2)$ \\
            \addlinespace[0.4em]
            $4,000$ & $100\%$ & $2,000$ & $1,000$ & $0.005$ & $150$ & $D_t U =  -  0.8162(D_x^3 U) -  0.4323(D_x U^2)$ \\
            \addlinespace[0.4em]
            $2,000$ & $50\%$ & $2,000$ & $1,000$ & $0.005$ & $100$ & $D_t U =  -  0.9767(D_x^3 U) -  0.4823(D_x U^2)$ \\
            \addlinespace[0.4em]
            $1,000$ & $50\%$ & $2,000$ & $1,000$ & $0.005$ & $100$ & $D_t U =  -  1.0205(D_x^3 U) -  0.4763(D_x U^2)$ \\
            \addlinespace[0.4em]
            $500$ & $50\%$ & $2,000$ & $1,000$ & $0.005$ & $50$ & $D_t U =  -  0.7997(D_x^3 U) -  0.4057(D_x U^2)$ \\
            \bottomrule[0.3ex]
        \end{tabulary}
    \end{threeparttable}
\end{table}

\bigskip 

Table \ref{Table:Experiments:KdV} shows that \texttt{weak-PDE-LEARN} successfully identifies the KdV equation in all experiments. 
It can learn the KdV equation with as few as $500$ data points, even when we contaminate that data with $50\%$ noise.
In most experiments, the coefficients in the learned PDE are close to those in the hidden PDE.
As we observed with Burgers' equation, however, the accuracy of the learned coefficients decreases as the noise level increases, or as the number of data points decreases.

\bigskip 

\begin{figure}
\centering
\begin{subfigure}{.49\linewidth}
  \centering
  \includegraphics[width=\linewidth]{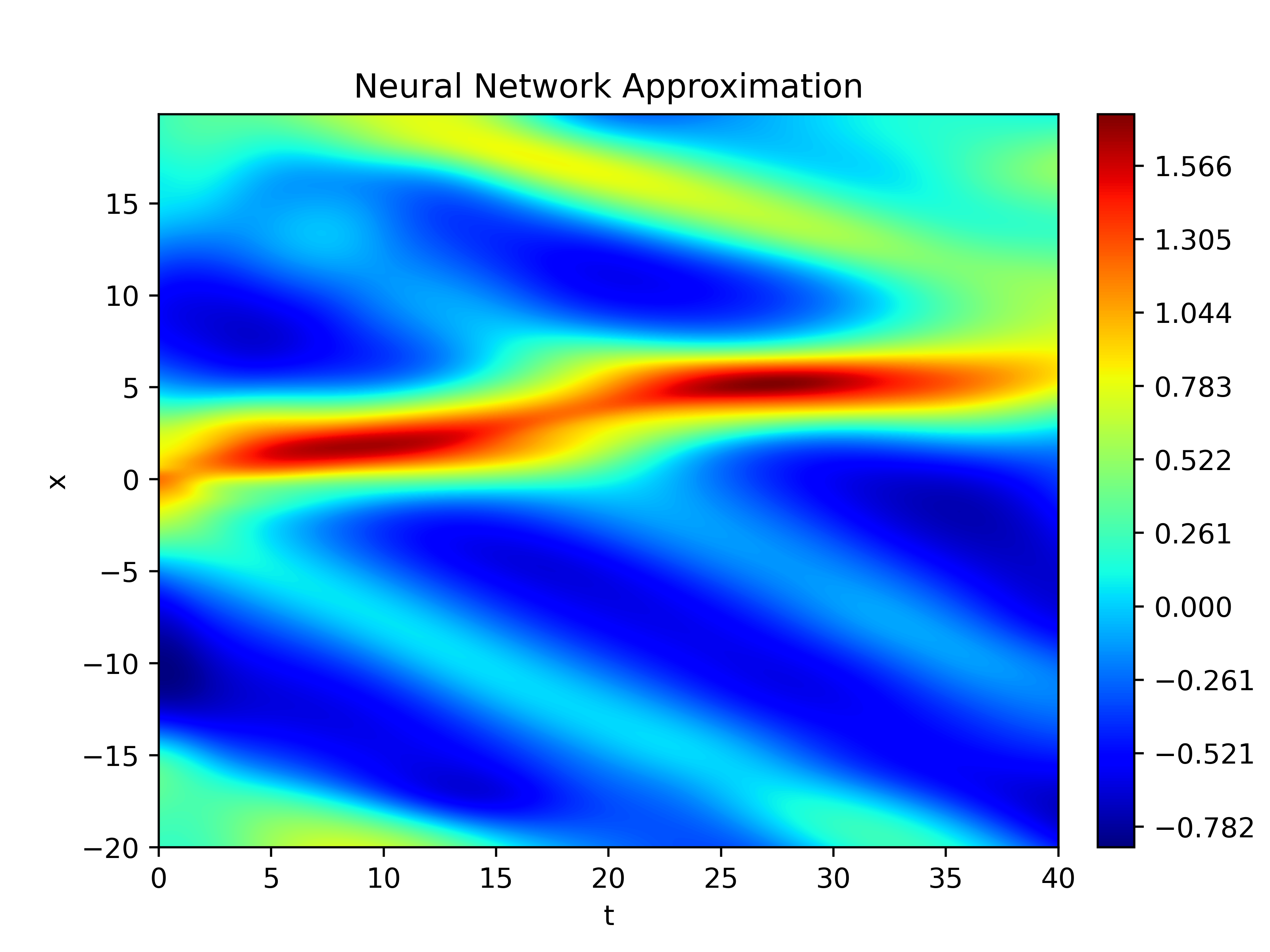}
  \caption{$U$ in the $25$\% noise, $4,000$ data point experiment.}
  \label{SubFig:Experiments:KdV:U25}
\end{subfigure}
\begin{subfigure}{.49\linewidth}
  \centering
  \includegraphics[width=\linewidth]{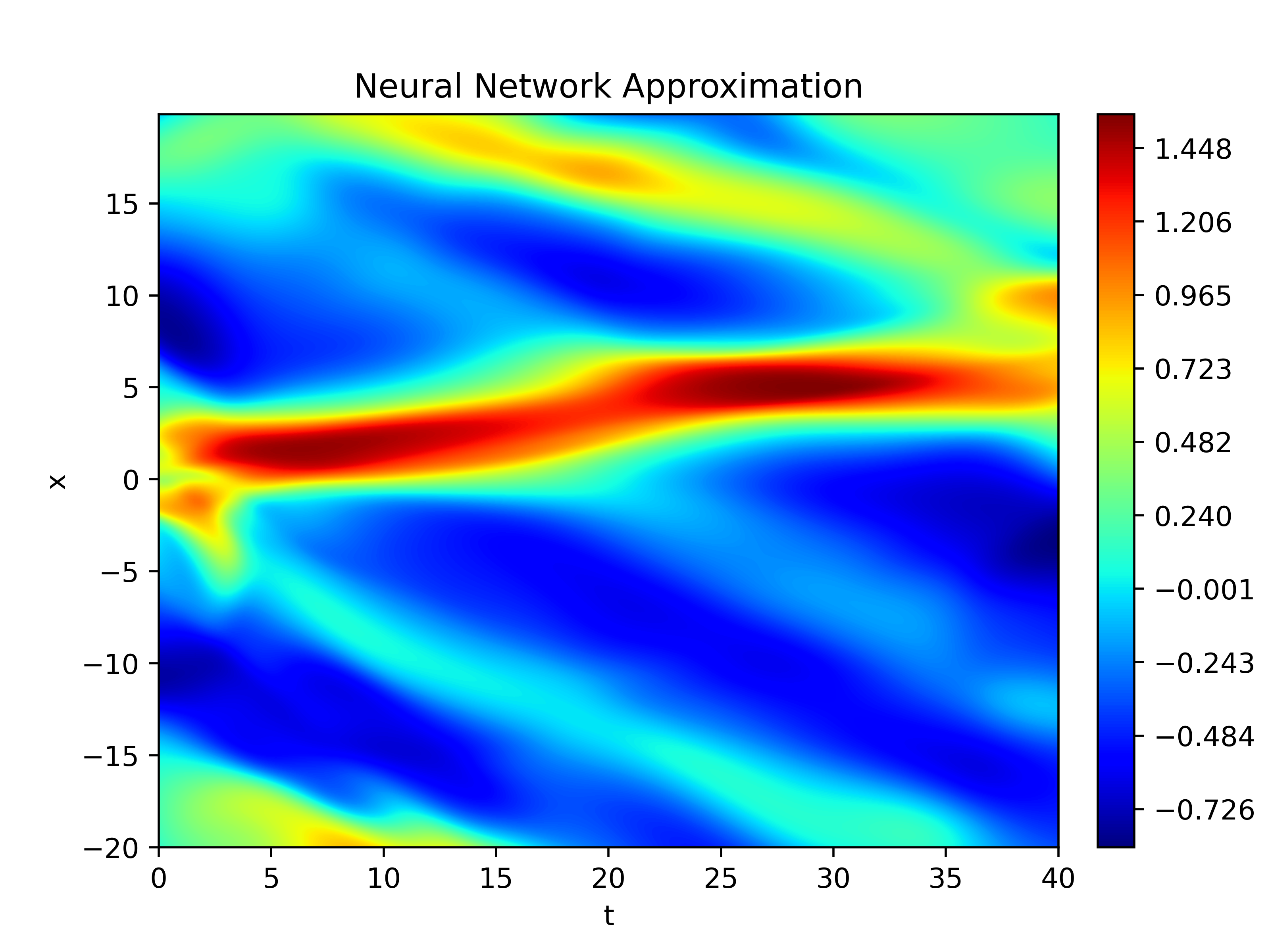}
  \caption{$U$ in the $75$\% noise, $4,000$ data point experiment.}
  \label{SubFig:Experiments:KdV:U75}
\end{subfigure}
\caption{Comparison of $U$ in the $25$\% and $75$\% noise, $4,000$ data point KdV equation experiments}
\label{Fig:Experiments:KdV:U}
\end{figure}

Figure \ref{Fig:Experiments:KdV:U} depicts $U$ from the $25$\% and $75$\% noise experiments (with $4,000$ data points).
Both plots closely resemble the noise-free data set.
As with Burgers' equation, there are some differences between $U$ and the system response function.
These differences are likely the result of the fact that we train $U$ to satisfy the weak form of the hidden PDE.
This result demonstrates that \texttt{weak-PDE-LEARN} can recover the system response function, even from data sets with high noise levels. 

\begin{figure}[!hbt]
    \centering
    \includegraphics[width=.8\linewidth]{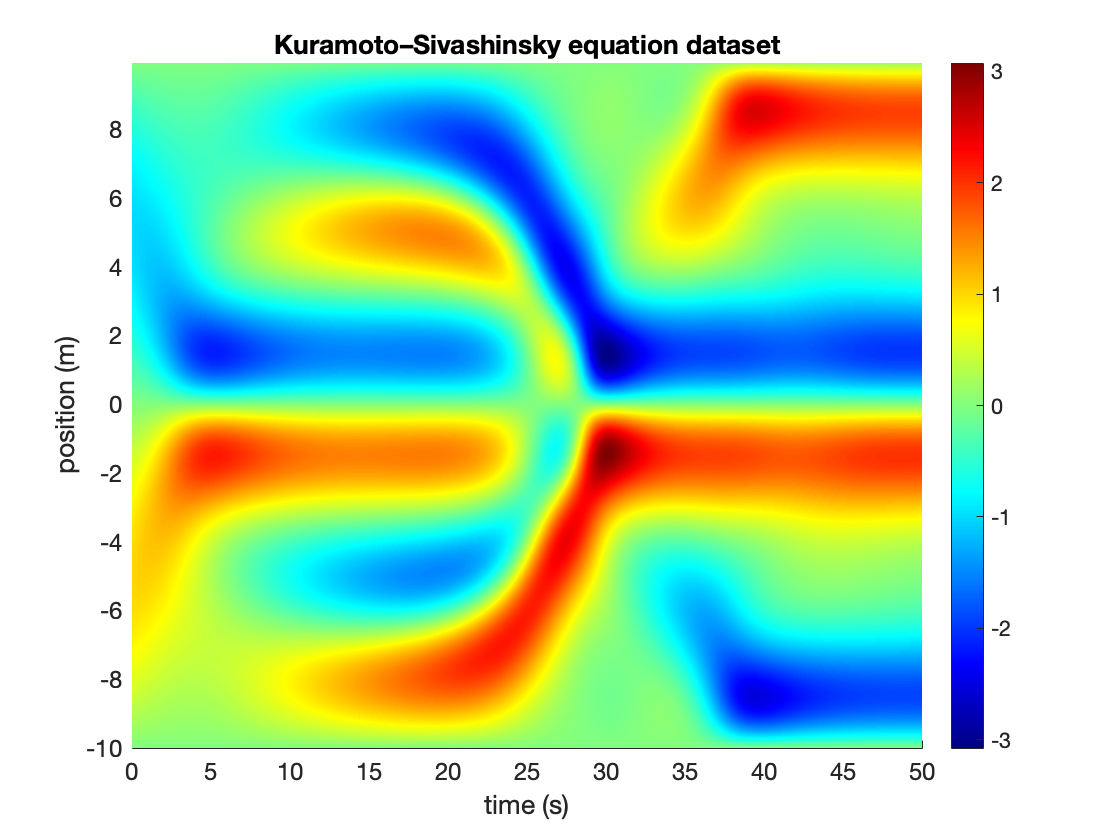}
    \caption{Noise-free KS equation data set.}
    \label{Fig:Experiments:KS:Dataset}
\end{figure}

\subsection{Kuromoto-Shivashinky Equation}
\label{SubSec:Experiments:KS}

The Kuroromoto-Shivashinsky (KS) equation is a non-linear fourth-order PDE \cite{kuramoto1976persistent} \cite{sivashinsky1977nonlinear}.
The KS equation arises in many physical contexts, including flame propagation, plasma physics, chemical physics, and combustion dynamics \cite{papageorgiou1991route}.
The KS equation is 

$$D_t u = \nu  D_x^2 u - \mu  D_x^4 u - 0.5 \lambda  D_x u^2,$$

where $\nu$, $\mu$, and $\lambda$ are constants.
If $\nu < 0$, solutions to the KS equation can be chaotic and develop violent shocks \cite{kuramoto1976persistent} \cite{papageorgiou1991route}.

\bigskip 

We test \texttt{weak-PDE-LEARN} on the KS equation with $\nu = -1$, $\mu = 1$, and $\lambda = 1$.
We use \texttt{Chebfun} to generate noisy, limited data sets.
For these experiments, $[0, T] \times \Omega = [0, 50] \times [-10, 10]$ and 

$$u(0, x) = -\sin\left(\pi \frac{x}{10}\right).$$

The script \texttt{KS\_Sine.m} in the \texttt{MATLAB} sub-directory of our repository uses \texttt{Chebfun}'s \texttt{spin} class to generate the noise-free data set.
Figure \ref{Fig:Experiments:KS:Dataset} depicts the noise-free data set.
We use the procedure outlined at the start of this section to make a collection of noisy and limited data sets.
We then test \texttt{weak-PDE-LEARN} on each data set.
Table \ref{Table:Experiments:KS} reports the results of these experiments.

\begin{table}[hbt]
    \centering 
    \rowcolors{2}{white}{cyan!10}
    \begin{threeparttable}
        \caption{Experimental results for KS equation} 
        \label{Table:Experiments:KS}

        \begin{tabulary}{\linewidth}{p{1.1cm}p{0.9cm}p{1.0cm}p{1.2cm}p{1.0cm}p{1.0cm}L}
            \toprule[0.3ex]
            $N_{Data}$ & Noise & $N_{Burn}$ & $N_{Sparse}$ & $\lambda_{L^p}$ & $N_{Tune}$ & Identified PDE \\
            \midrule[0.1ex] 
            $10,000$ & $25\%$ & $2,000$ & $2,000$ & $0.005$ & $100$ & $D_t U =  -0.99(D_x^2 U) -  0.99(D_x^4 U) -  0.50(D_x U^2)$ \\
            \addlinespace[0.4em]
            $10,000$ & $50\%$ & $2,000$ & $2,000$ & $0.005$ & $100$ & $D_t U =  -0.96(D_x^2 U) -  0.96(D_x^4 U) -  0.49(D_x U^2)$ \\
            \addlinespace[0.4em]
            $10,000$ & $75\%$ & $2,000$ & $3,000$ & $0.005$ & $100$ & $D_t U =  -0.89(D_x^2 U) -  0.87(D_x^4 U) -  0.47(D_x U^2)$ \\
            \addlinespace[0.4em]
            $10,000$ & $100\%$ & $2,000$ & $2,000$ & $0.005$ & $150$ & $D_t U =  0.15(U) -  0.18(D_x^4 U) -  0.40(D_x U^2) -  0.12(D_x^3 U^2)$ \\
            \addlinespace[0.4em]
            $4,000$ & $50\%$ & $2,000$ & $1,000$ & $0.005$ & $100$ & $D_t U =  -1.09(D_x^2 U) -  1.08(D_x^4 U) -  0.55(D_x U^2)$ \\
            \addlinespace[0.4em]
            $2,000$ & $50\%$ & $2,000$ & $1,000$ & $0.005$ & $100$ & $D_t U =  -0.83(D_x^2 U) -  0.84(D_x^4 U) -  0.45(D_x U^2)$ \\
            \addlinespace[0.4em]
            $1,000$ & $50\%$ & $2,000$ & $1,000$ & $0.005$ & $50$ & $D_t U =  -0.57(D_x^2 U) -  0.55(D_x^4 U) -  0.37(D_x U^2)$ \\
            \addlinespace[0.4em]
            $500$ & $25\%$ & $2,000$ & $1,000$ & $0.005$ & $50$ & $D_t U =  -0.67(D_x^2 U) -  0.61(D_x^4 U) -  0.38(D_x U^2)$ \\
            \addlinespace[0.4em]
            $500$ & $10\%$ & $2,000$ & $1,000$ & $0.005$ & $50$ & $D_t U =  -0.92(D_x^2 U) -  0.91(D_x^4 U) -  0.47(D_x U^2)$ \\
            \bottomrule[0.3ex]
        \end{tabulary}
    \end{threeparttable}
\end{table}

\bigskip 

We observe that \texttt{weak-PDE-LEARN} identifies the KS equation in all but one experiment.
In particular, \texttt{weak-PDE-LEARN} fails to learn the KS equation in the $100\%$ noise, $10,000$ data point experiment. 
Notably, even when it fails, the identified PDE contains two terms ($D_x^4 U$ and $D_x U^2$) that appear in the true PDE.
This result suggests that \texttt{weak-PDE-LEARN} can discover information about the underlying dynamics even when it can not fully identify the governing PDE. 

\bigskip 

It is worth considering the failed experiment in more detail.
\texttt{Weak-PDE-LEARN} correctly identifies the KdV and Burgers' equation equations at this noise level.
However, the fourth-order derivatives of the KS equation and the relatively intricate nature of the KS data set may make identifying the KS equation more challenging than identifying the Burgers' or KdV equations.
Notably, \texttt{weak-PDE-LEARN} identifies the KS equation in data sets with a lower noise level, even with considerably fewer data points (for example, the $50$\% noise, $1,000$ data point experiment).
This result may suggest that extreme noise is more of a limiting factor than data set size. 

\bigskip 

\begin{figure}
\centering
\begin{subfigure}{.49\linewidth}
  \centering
  \includegraphics[width=\linewidth]{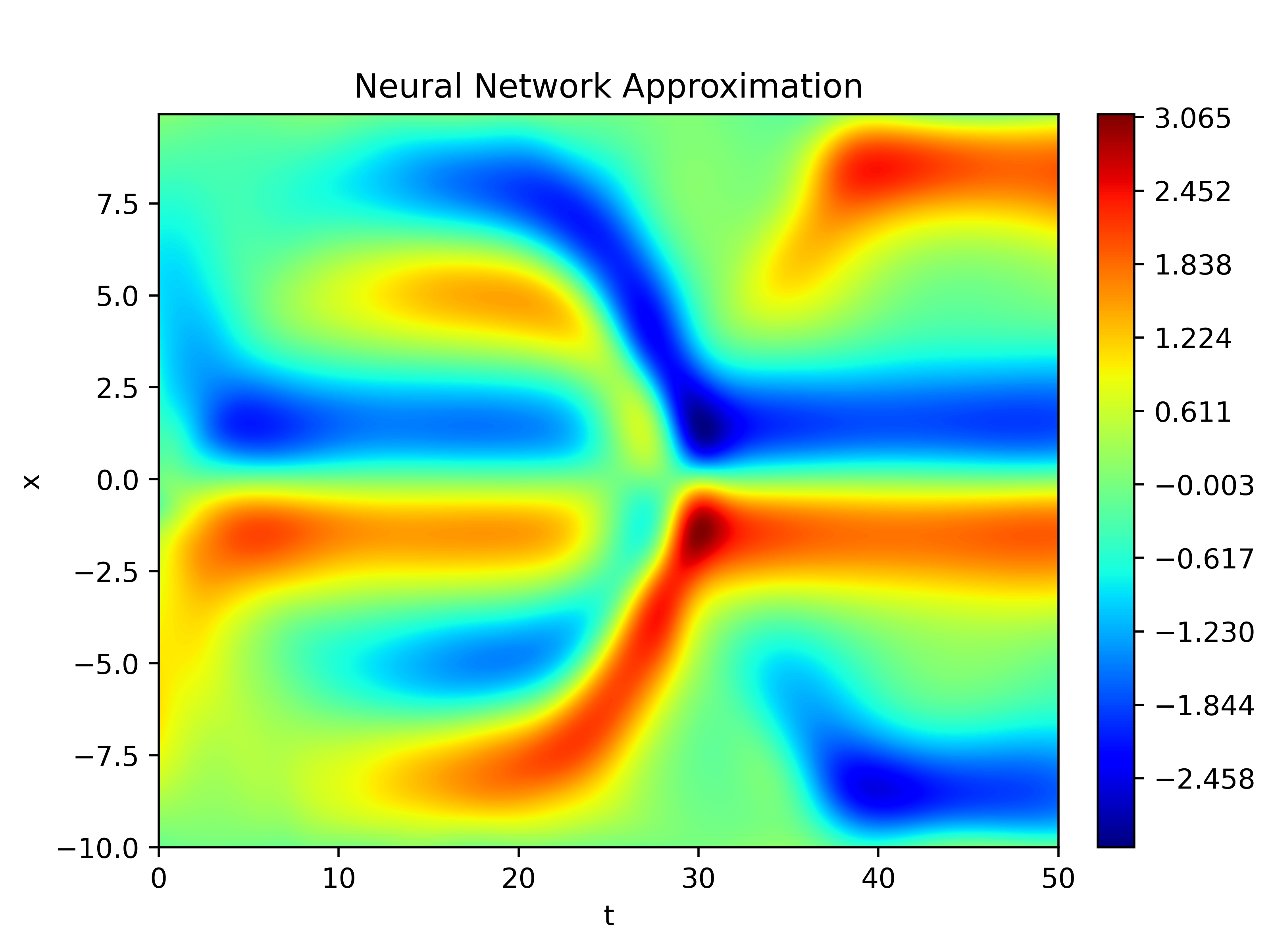}
  \caption{$U$ in the $25$\% noise, $10,000$ data point experiment.}
  \label{SubFig:Experiments:KS:U25}
\end{subfigure}
\begin{subfigure}{.49\linewidth}
  \centering
  \includegraphics[width=\linewidth]{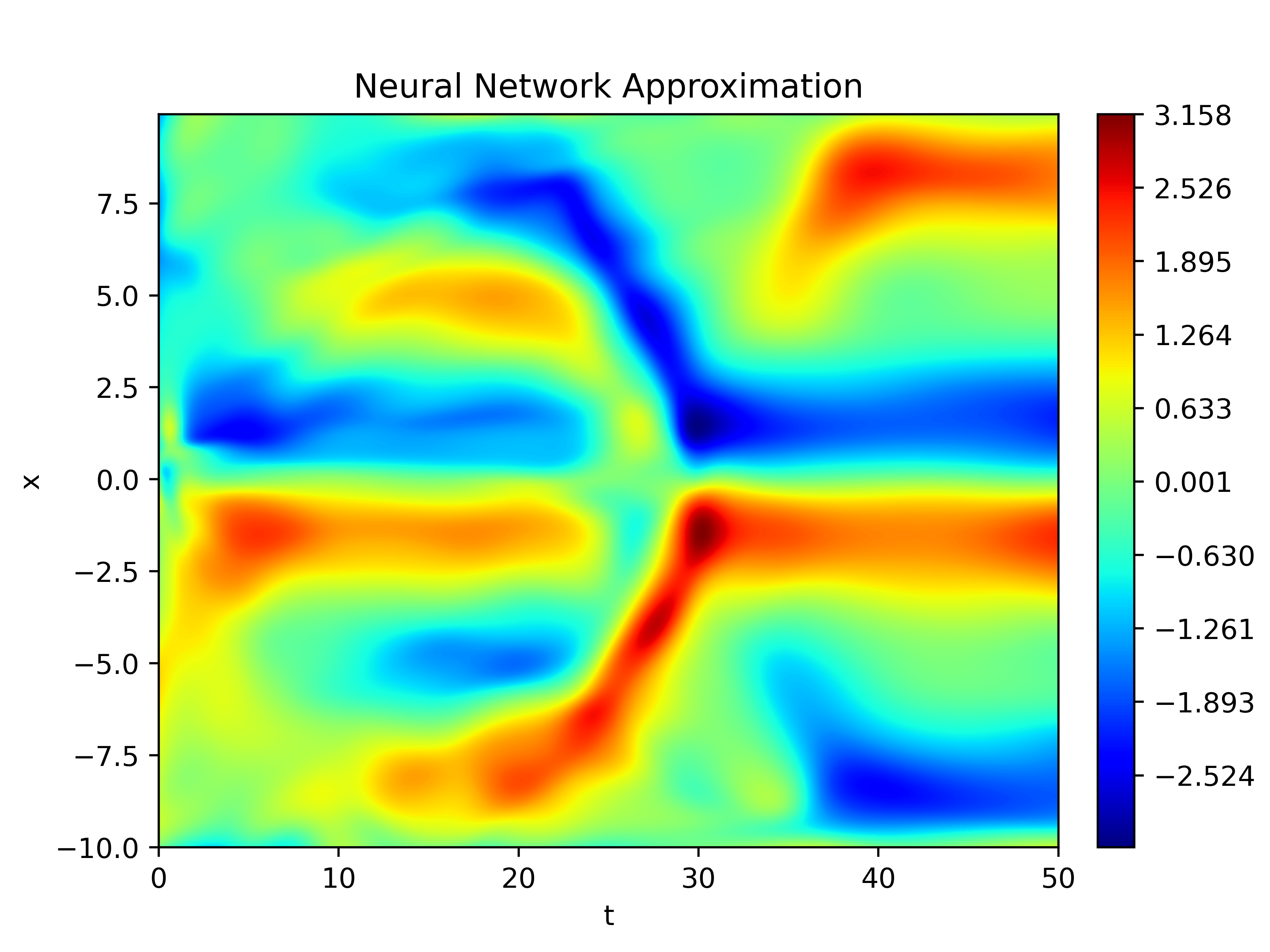}
  \caption{$U$ in the $75$\% noise, $10,000$ data point experiment.}
  \label{SubFig:Experiments:KS:U75}
\end{subfigure}
\caption{Comparison of $U$ in the $25$\% and $75$\% noise, $10,000$ data point KS equation experiments}
\label{Fig:Experiments:KS:U}
\end{figure}

Figure \ref{Fig:Experiments:KS:U} depicts the learned solution ($U$) from the $25$\% and $75$\% noise, $10,000$ data point experiments.
Both plots closely resemble the noise-free data set, Figure \ref{Fig:Experiments:KS:Dataset}.

\section{Discussion}
\label{Sec:Discussion}

In this section, we discuss further aspects of the \texttt{weak-PDE-LEARN} algorithm. 
First, in section \ref{SubSec:Discussion:Weak}, we discuss our rationale for developing a PDE-discovery algorithm built around the weak form of the hidden PDE.
Then, in section \ref{SubSec:Discussion:Limitations}, we discuss some key limitations of this approach.

\subsection{Weak Forms and noise}
\label{SubSec:Discussion:Weak}

The weak-form approach outlined in section \ref{SubSec:Method:weak_PDE_LEARN} forms the core of the \texttt{weak-PDE-LEARN} algorithm. 
The results of the preceding section demonstrate that this approach yields an effective tool for identifying PDEs from noisy and limited data sets.
We chose this approach for two main reasons, which we will now discuss. 

\bigskip 

First, it allowed us to construct a loss function, $\text{Loss}_{\text{weak}}$, that enforces the hidden PDE but does not depend on the partial derivatives of $U$.
We can not overstate the significance of this fact.
The hidden PDE, equation \eqref{Eq:PDE}, depends on $u$ and its partial derivatives. 
We also know that noise tends to be high-frequency.
Differentiation is an unbounded operator that amplifies high-frequency signals \cite{kreyszig1991introductory}. 
Thus, differentiation tends to amplify noise.
The original \texttt{PDE-LEARN} algorithm \cite{stephany2022LEARN} enforces the hidden PDE at a set of \emph{collocation points}.
This approach requires the partial derivatives of $U$, which the algorithm computes using automatic differentiation.
If $U$ picks up any noise from the underlying data set, the partial derivatives of $U$ can differ considerably from those of the system response function, $u$. 
This unfortunate result may explain why \texttt{PDE-LEARN} has trouble learning the fourth-order KS equation (but can learn lower-order equations with relative ease).

\bigskip 

Second, a consequence of the definition of the Riemannian integral is the following: Given an interval, $[a, b]$, a Riemann integrable function, $f$, and an $\varepsilon > 0$, there exists a partition, $P = \{x_0, \ldots, x_N \}$ of $[a, b]$ such that if $t_i \in [x_{i - 1}, x_{i}]$, the weighted average 
$$\sum_{i = 1}^{N} f(t_i) (x_{i} - x_{i - 1})$$ 
is within $\varepsilon$ of $\int_{a}^{b} f$ \cite{rudin1976principles}.
In other words, the Riemann integral of a function acts like a scaled average of that function over the integration domain. 
In our case, the function we are integrating has additive noise with a mean of zero.
Thus, by additivity of the integral, we expect the integral of a function with added noise to roughly match that of just the function (as the integral of the noise has zero expected value).
Based on this, it is reasonable to expect that using integration to enforce the hidden PDE will help \texttt{weak-PDE-LEARN} perform well on data with a mean of zero.
Section \ref{Sec:Experiments} confirms those expectations.
Notably, this argument breaks down if the noise does not have a mean of zero, which could conceivably happen in practice.
A potential future research direction is exploring how \texttt{weak-PDE-LEARN} performs on data with other noise models.

\subsection{Limitations}
\label{SubSec:Discussion:Limitations}

PDE discovery algorithms build around the weak form of the hidden PDE have some limitations.
To arrive at equation \eqref{Eq:PDE:Weak}, we used Green's Lemma to move the derivatives from the library terms to the weight functions.
This approach only works if each library term has the form $D^{\alpha} f(u)$, for some multi-index $\alpha$. 
Thus, if the hidden PDE does not have the form of equation \eqref{Eq:PDE}, then 
\texttt{weak-PDE-LEARN} can not discover it. 
To realize the advantages of using the weak form of the hidden PDE, we must place additional constraints on the hidden PDE.

\bigskip 

Notably, even if the hidden PDE does not have the form of equation \eqref{Eq:PDE}, we can still use Green's Lemma to reduce the number of partial derivatives of $U$ that we need to compute.
For example, consider the library term $(D^2_x U)(D_x U) = D_x (D_x U^2)$.
While we can not use Green's lemma to offload all of $U$'s derivatives onto the weight function, we do have 
$$\int_{[0, T] \times \Omega} w(t, X) \Big( \big(D^2_x U\big)\big(D_x U \big) \Big) \ dt\ d\Omega = -\int_{[0, T] \times \Omega} \Big( D_x w(t, X) \Big) \left( D_x U \right)^2\ dt\ d\Omega.$$
The expression on the right depends only on $D_x U$, while the one on the left depends on $D_x U$ and $D_x^2 U$. 
Thus, we no longer need to compute the second partial derivative of $U$ to evaluate the integral on the left. 
Nonetheless, the advantages of using a form based around the weak form of the hidden PDE are strongest when that PDE has the form of equation \eqref{Eq:PDE}.

\bigskip 

There are several existing PDE discovery algorithms (such as \texttt{PDE-LEARN} \cite{stephany2022LEARN}, \texttt{PDE-READ} \cite{stephany2022LEARN}, \texttt{DeepMoD} \cite{both2021deepmod}, or \cite{chen2021physics}) that do not suffer from this limitation.
Each of these approaches incorporates the hidden PDE (in the form of equation \eqref{Eq:PDE}) directly into their loss functions (\emph{e.g.},  the collocation loss in \texttt{PDE-LEARN} \cite{stephany2022LEARN}).
We refer to algorithms that take this approach as \emph{strong form approaches}.
Because of their favorable performance with noisy data, PDE discovery algorithms built around the weak form of the hidden PDE, such as \texttt{weak-PDE-LEARN}, may be the best choice when it is reasonable to assume the hidden PDE has the form of equation \eqref{Eq:PDE}.
Otherwise, strong-form approaches may be a better choice.

\bigskip 

We conclude this discussion with an interesting further limitation of using the weak form of the hidden PDE.
Above, we argued an approach built around integration should perform well with noisy data.
In practice, however, \texttt{weak-PDE-LEARN} doesn't integrate the system response function, $u$.
Instead, it integrates an approximation of $u$. 
Since integration tends to average out small features, learning fine (small length-scale) features in the data set may not significantly reduce $\text{Loss}_{\text{weak}}$.
This result means that if learning fine features in a data set is critical to identifying the hidden PDE, then \texttt{weak-PDE-LEARN} may struggle to identify the hidden PDE.

\bigskip 

\begin{figure}[!hbt]
    \centering
    \includegraphics[width=.8\linewidth]{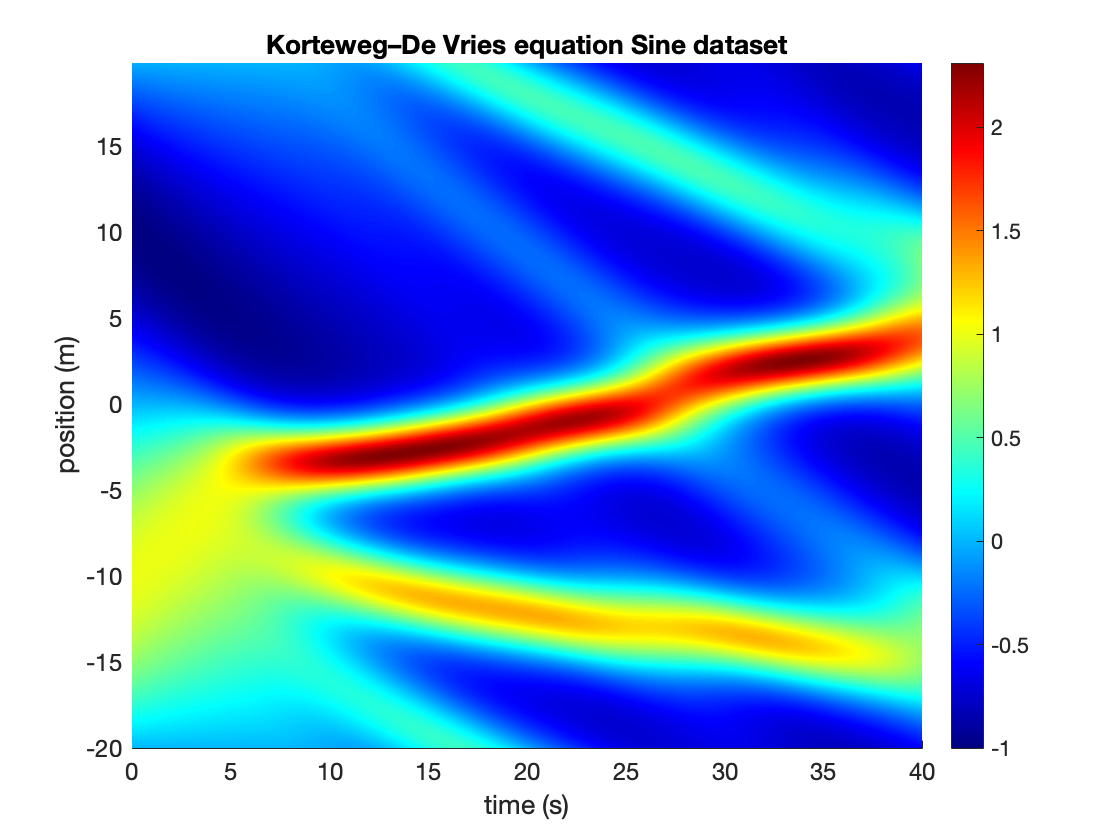}
    \caption{Noise-free KdV-Sine' equation data set.}
    \label{Fig:Experiments:KdV_Sine:Dataset}
\end{figure}

This limitation is not purely theoretical, either.
While testing with the \texttt{KdV} equation, we tried learning the KdV equation from a second data set. 
In this data set, $[0, T] \times \Omega = [0, 40] \times [-10, 10]$ and $u(0, x) = -\sin(\pi x / 10)$. 
We refer to this as the \emph{KdV-Sine} data set.
Figure \ref{Fig:Experiments:KdV_Sine:Dataset} depicts the noise-free data set.

\bigskip 

\begin{figure}
\centering
\begin{subfigure}{.49\linewidth}
  \centering
  \includegraphics[width=\linewidth]{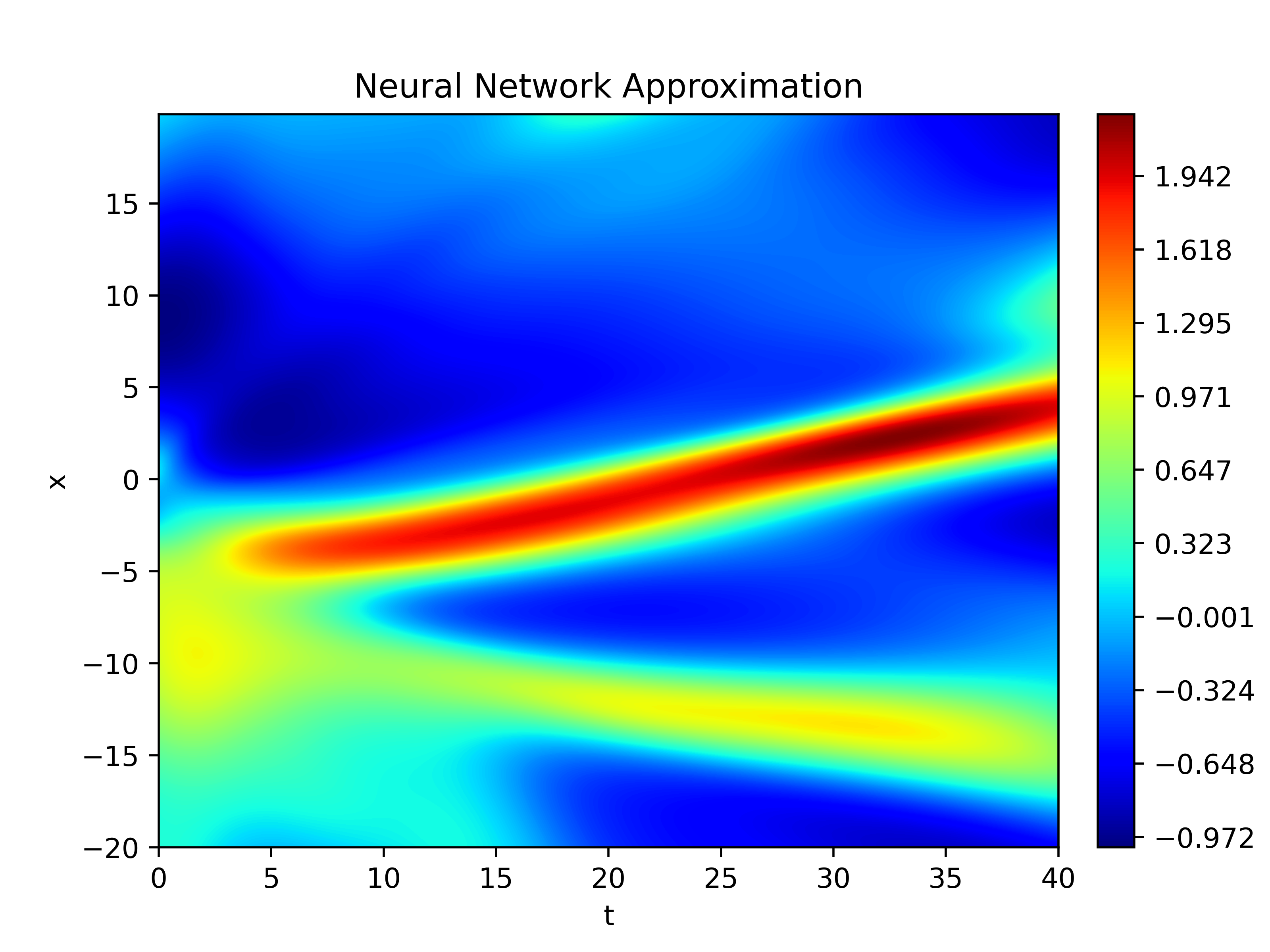}
  \caption{$U$ in the $25$\% noise, $4,000$ data point experiment.}
  \label{SubFig:Experiments:KdV_Sine:Adam}
\end{subfigure}
\begin{subfigure}{.49\linewidth}
  \centering
  \includegraphics[width=\linewidth]{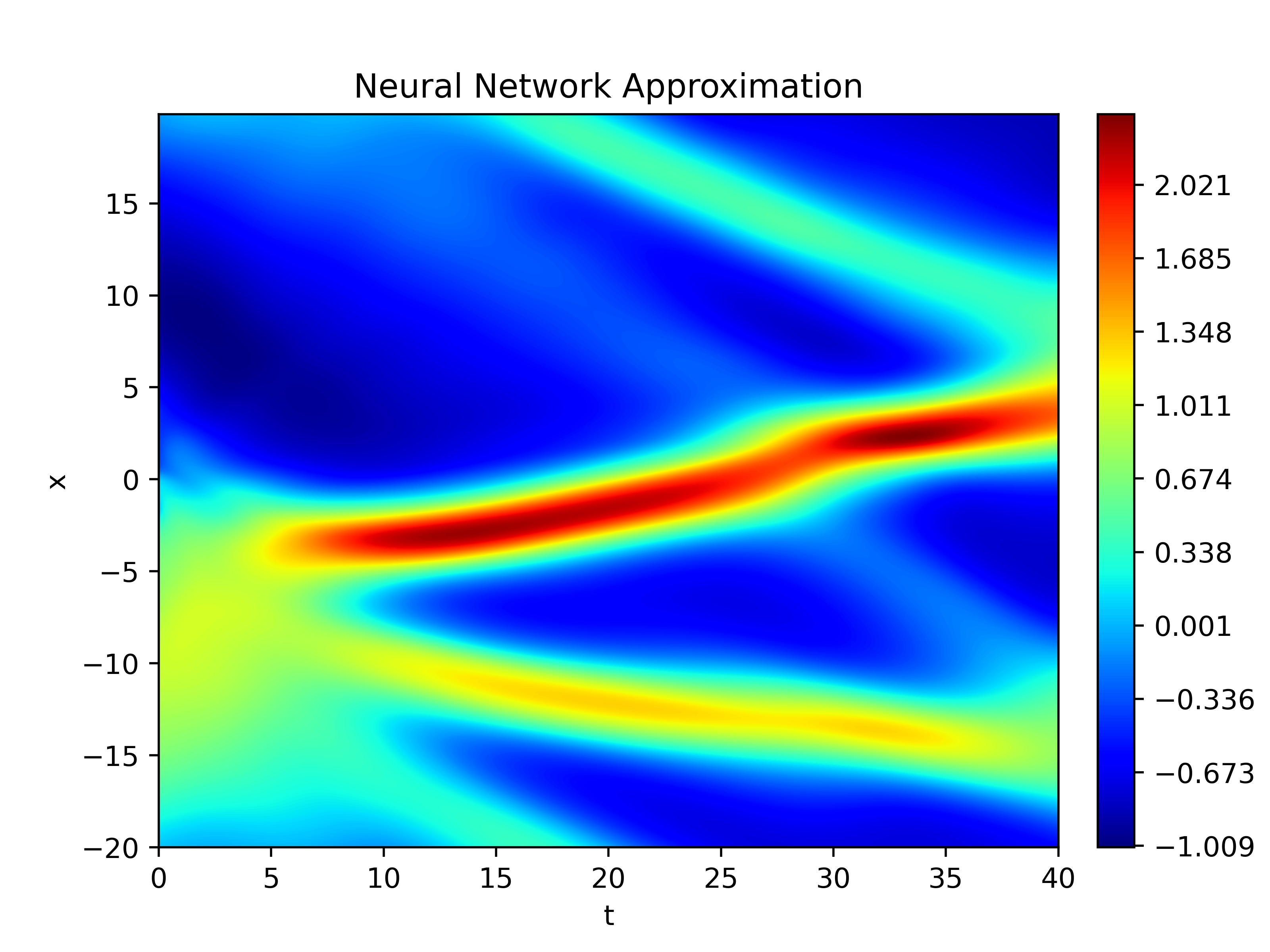}
  \caption{$U$ after training in the $25$\% noise, $4,000$ data point experiment experiment.}
  \label{SubFig:Experiments:KdV_Sine:LBFGS}
\end{subfigure}
\caption{Comparison of $U$ in the $25$\% and $75$\% noise, $4,000$ data point KdV-Sine equation experiments}
\label{Fig:Experiments:KdV_Sine:U}
\end{figure}

We tested \texttt{weak-PDE-LEARN} on this data set.
With $4,000$ data points and $25$\% noise, it successfully identifies the KdV equation, though training takes far longer than in our experiments in section \ref{Sec:Experiments}.
In particular, \texttt{weak-PDE-LEARN} needed $6,000$ burn-in epochs and $5,000$ sparsification epochs. 
One key difference between the KdV-Sine data set and the one we considered in subsection \ref{SubSec:Experiments:KdV} is that the former exhibits several subtle response features.
\texttt{Weak-PDE-LEARN} appears to learn these features very slowly.
To demonstrate this, Figure \ref{SubFig:Experiments:KdV_Sine:Adam} depicts $U$ after $4,000$ burn-in epochs.
We can see that $U$ has picked up the major features of Figure \ref{Fig:Experiments:KdV_Sine:Dataset}, such as the large wavefront across the middle of the problem domain, but not some of the more subtle ones, such as the smaller and less pronounced wavefronts.
Figure \ref{SubFig:Experiments:KdV_Sine:LBFGS}, which closely resembles Figure \ref{Fig:Experiments:KdV_Sine:Dataset}, depicts $U$ after completing training.
Thus, \texttt{weak-PDE-LEARN} eventually learns the fine features of the KdV-Sine data set; it is just slow. 

\bigskip 

\section{Conclusion}
\label{Sec:Conclusion}

In this paper, we introduced \texttt{weak-PDE-LEARN}, a PDE discovery algorithm that learns a hidden PDE directly from noisy, limited measurements reminiscent of the data we can expect from physical experiments.
\texttt{Weak-PDE-LEARN} trains a rational neural network, $U$, to learn an approximation to a system response function, $u$, using noisy and limited measurements.
Simultaneously, it identifies a hidden PDE that $u$ satisfies. 
\texttt{Weak-PDE-LEARN} accomplishes these tasks by building the weak form of the hidden PDE into its loss function. 
This loss regularizes $U$, helping it de-noise the noisy measurements and learn an accurate approximation for $u$.
Notably, assuming the hidden PDE takes a particular form (see equation \eqref{Eq:PDE}) and using specially designed weight functions, our approach does not require us to evaluate the partial derivatives of $U$.
This feature helps \texttt{weak-PDE-LEARN} mitigate the deleterious effects of noise contamination. 
Additionally, we introduced an adaptive procedure for selecting weight functions, which further facilities \texttt{weak-PDE-LEARN}'s ability to identify a governing PDE that describes the system of interest.
As our experiments demonstrate, \texttt{weak-PDE-LEARN} is an effective tool for PDE discovery that can identify non-linear PDEs from noisy, limited measurements of their solutions.

\section{Acknowledgements}
The authors thank Maria Oprea for helpful discussions and advice on weak forms while designing \texttt{weak-PDE-LEARN}.
She first suggested we use weight functions based on the classic `bump function' and gave us numerous ideas to improve our algorithm.
Further, this work is supported by the Office of Naval Research (ONR) under grant N00014-22-1-2055.
Finally, Robert Stephany is supported by his NDSEG fellowship.

\printbibliography 

@article{abreu2019mortality,
  title={Mortality causes universal changes in microbial community composition},
  author={Abreu, Clare I and Friedman, Jonathan and Woltz, Vilhelm L Andersen and Gore, Jeff},
  journal={Nature communications},
  volume={10},
  number={1},
  pages={1--9},
  year={2019},
  publisher={Nature Publishing Group}
}

@article{amor2020transient,
  title={Transient invaders can induce shifts between alternative stable states of microbial communities},
  author={Amor, Daniel R and Ratzke, Christoph and Gore, Jeff},
  journal={Science advances},
  volume={6},
  number={8},
  pages={eaay8676},
  year={2020},
  publisher={American Association for the Advancement of Science}
}

@article{boulle2022data,
  title={Data-driven discovery of Green’s functions with human-understandable deep learning},
  author={Boull{\'e}, Nicolas and Earls, Christopher J and Townsend, Alex},
  journal={Scientific reports},
  volume={12},
  number={1},
  pages={4824},
  year={2022},
  publisher={Nature Publishing Group UK London}
}

@article{gin2021deepgreen,
  title={DeepGreen: deep learning of Green’s functions for nonlinear boundary value problems},
  author={Gin, Craig R and Shea, Daniel E and Brunton, Steven L and Kutz, J Nathan},
  journal={Scientific reports},
  volume={11},
  number={1},
  pages={21614},
  year={2021},
  publisher={Nature Publishing Group UK London}
}

@article{greydanus2019hamiltonian,
  title={Hamiltonian neural networks},
  author={Greydanus, Samuel and Dzamba, Misko and Yosinski, Jason},
  journal={Advances in neural information processing systems},
  volume={32},
  year={2019}
}

@inproceedings{chartrand2008iteratively,
  title={Iteratively reweighted algorithms for compressive sensing},
  author={Chartrand, Rick and Yin, Wotao},
  booktitle={2008 IEEE international conference on acoustics, speech and signal processing},
  pages={3869--3872},
  year={2008},
  organization={IEEE}
}

@article{kingma2014adam,
  title={Adam: A method for stochastic optimization},
  author={Kingma, Diederik P and Ba, Jimmy},
  journal={arXiv preprint arXiv:1412.6980},
  year={2014}
}

@article{liu1989limited,
  title={On the limited memory BFGS method for large scale optimization},
  author={Liu, Dong C and Nocedal, Jorge},
  journal={Mathematical programming},
  volume={45},
  number={1},
  pages={503--528},
  year={1989},
  publisher={Springer}
}

@article{stephany2022LEARN,
  title={PDE-LEARN: Using Deep Learning to Discover Partial Differential Equations from Noisy, Limited Data},
  author={Stephany, Robert and Earls, Christopher},
  journal={arXiv preprint arXiv:2212.04971},
  year={2022}
}

@article{stephany2022READ,
  title={PDE-READ: Human-readable partial differential equation discovery using deep learning},
  author={Stephany, Robert and Earls, Christopher},
  journal={Neural Networks},
  volume={154},
  pages={360--382},
  year={2022},
  publisher={Elsevier}
}

@article{boulle2020rational,
  title={Rational neural networks},
  author={Boull{\'e}, Nicolas and Nakatsukasa, Yuji and Townsend, Alex},
  journal={Advances in Neural Information Processing Systems},
  volume={33},
  pages={14243--14253},
  year={2020}
}

@article{bongard2007automated,
  title={Automated reverse engineering of nonlinear dynamical systems},
  author={Bongard, Josh and Lipson, Hod},
  journal={Proceedings of the National Academy of Sciences},
  volume={104},
  number={24},
  pages={9943--9948},
  year={2007},
  publisher={National Acad Sciences}
}

@article{schmidt2009distilling,
  title={Distilling free-form natural laws from experimental data},
  author={Schmidt, Michael and Lipson, Hod},
  journal={science},
  volume={324},
  number={5923},
  pages={81--85},
  year={2009},
  publisher={American Association for the Advancement of Science}
}

@article{brunton2016discovering,
  title={Discovering governing equations from data by sparse identification of nonlinear dynamical systems},
  author={Brunton, Steven L and Proctor, Joshua L and Kutz, J Nathan},
  journal={Proceedings of the national academy of sciences},
  volume={113},
  number={15},
  pages={3932--3937},
  year={2016},
  publisher={National Acad Sciences}
}

@article{rudy2017data,
  title={Data-driven discovery of partial differential equations},
  author={Rudy, Samuel H and Brunton, Steven L and Proctor, Joshua L and Kutz, J Nathan},
  journal={Science Advances},
  volume={3},
  number={4},
  pages={e1602614},
  year={2017},
  publisher={American Association for the Advancement of Science}
}

@article{berg2019data,
  title={Data-driven discovery of PDEs in complex datasets},
  author={Berg, Jens and Nystr{\"o}m, Kaj},
  journal={Journal of Computational Physics},
  volume={384},
  pages={239--252},
  year={2019},
  publisher={Elsevier}
}

@article{schaeffer2017learning,
  title={Learning partial differential equations via data discovery and sparse optimization},
  author={Schaeffer, Hayden},
  journal={Proceedings of the Royal Society A: Mathematical, Physical and Engineering Sciences},
  volume={473},
  number={2197},
  pages={20160446},
  year={2017},
  publisher={The Royal Society Publishing}
}

@article{gurevich2019robust,
  title={Robust and optimal sparse regression for nonlinear PDE models},
  author={Gurevich, Daniel R and Reinbold, Patrick AK and Grigoriev, Roman O},
  journal={Chaos: An Interdisciplinary Journal of Nonlinear Science},
  volume={29},
  number={10},
  pages={103113},
  year={2019},
  publisher={AIP Publishing LLC}
}

@article{messenger2021weak,
  title={Weak SINDy for partial differential equations},
  author={Messenger, Daniel A and Bortz, David M},
  journal={Journal of Computational Physics},
  volume={443},
  pages={110525},
  year={2021},
  publisher={Elsevier}
}

@article{baydin2018automatic,
  title={Automatic differentiation in machine learning: a survey},
  author={Baydin, Atilim Gunes and Pearlmutter, Barak A and Radul, Alexey Andreyevich and Siskind, Jeffrey Mark},
  journal={Journal of machine learning research},
  volume={18},
  year={2018},
  publisher={Journal of Machine Learning Research}
}

@article{raissi2018deep,
  title={Deep hidden physics models: Deep learning of nonlinear partial differential equations},
  author={Raissi, Maziar},
  journal={The Journal of Machine Learning Research},
  volume={19},
  number={1},
  pages={932--955},
  year={2018},
  publisher={JMLR. org}
}

@article{atkinson2019data,
  title={Data-driven discovery of free-form governing differential equations},
  author={Atkinson, Steven and Subber, Waad and Wang, Liping and Khan, Genghis and Hawi, Philippe and Ghanem, Roger},
  journal={arXiv preprint arXiv:1910.05117},
  year={2019}
}

@article{bonneville2021bayesian,
  title={Bayesian Deep Learning for Partial Differential Equation Parameter Discovery with Sparse and Noisy Data},
  author={Bonneville, Christophe and Earls, Christopher J},
  journal={arXiv preprint arXiv:2108.04085},
  year={2021}
}

@article{guyon2002gene,
  title={Gene selection for cancer classification using support vector machines},
  author={Guyon, Isabelle and Weston, Jason and Barnhill, Stephen and Vapnik, Vladimir},
  journal={Machine learning},
  volume={46},
  number={1},
  pages={389--422},
  year={2002},
  publisher={Springer}
}

@article{paszke2019pytorch,
  title={Pytorch: An imperative style, high-performance deep learning library},
  author={Paszke, Adam and Gross, Sam and Massa, Francisco and Lerer, Adam and Bradbury, James and Chanan, Gregory and Killeen, Trevor and Lin, Zeming and Gimelshein, Natalia and Antiga, Luca and others},
  journal={Advances in neural information processing systems},
  volume={32},
  year={2019}
}

@article{paszke2017automatic,
  title={Automatic differentiation in pytorch},
  author={Paszke, Adam and Gross, Sam and Chintala, Soumith and Chanan, Gregory and Yang, Edward and DeVito, Zachary and Lin, Zeming and Desmaison, Alban and Antiga, Luca and Lerer, Adam},
  year={2017}
}

@article{basdevant1986spectral,
  title={Spectral and finite difference solutions of the Burgers equation},
  author={Basdevant, Cea and Deville, M and Haldenwang, P and Lacroix, JM and Ouazzani, J and Peyret, R and Orlandi, Paolo and Patera, AT},
  journal={Computers \& fluids},
  volume={14},
  number={1},
  pages={23--41},
  year={1986},
  publisher={Elsevier}
}

@article{bateman1915some,
  title={Some recent researches on the motion of fluids},
  author={Bateman, Harry},
  journal={Monthly Weather Review},
  volume={43},
  number={4},
  pages={163--170},
  year={1915}
}

@article{korteweg1895xli,
  title={XLI. On the change of form of long waves advancing in a rectangular canal, and on a new type of long stationary waves},
  author={Korteweg, Diederik Johannes and De Vries, Gustav},
  journal={The London, Edinburgh, and Dublin Philosophical Magazine and Journal of Science},
  volume={39},
  number={240},
  pages={422--443},
  year={1895},
  publisher={Taylor \& Francis}
}

@misc{driscoll2014chebfun,
  title={Chebfun guide},
  author={Driscoll, Tobin A and Hale, Nicholas and Trefethen, Lloyd N},
  year={2014},
  publisher={Pafnuty Publications, Oxford}
}

@article{sivashinsky1977nonlinear,
  title={Nonlinear analysis of hydrodynamic instability in laminar flames—I. Derivation of basic equations},
  author={Sivashinsky, Gregory I},
  journal={Acta astronautica},
  volume={4},
  number={11},
  pages={1177--1206},
  year={1977}
}

@article{kuramoto1976persistent,
  title={Persistent propagation of concentration waves in dissipative media far from thermal equilibrium},
  author={Kuramoto, Yoshiki and Tsuzuki, Toshio},
  journal={Progress of theoretical physics},
  volume={55},
  number={2},
  pages={356--369},
  year={1976},
  publisher={Oxford University Press}
}

@article{papageorgiou1991route,
  title={The route to chaos for the Kuramoto-Sivashinsky equation},
  author={Papageorgiou, Demetrios T and Smyrlis, Yiorgos S},
  journal={Theoretical and Computational Fluid Dynamics},
  volume={3},
  number={1},
  pages={15--42},
  year={1991},
  publisher={Springer}
}

@article{both2021deepmod,
  title={DeepMoD: Deep learning for Model Discovery in noisy data},
  author={Both, Gert-Jan and Choudhury, Subham and Sens, Pierre and Kusters, Remy},
  journal={Journal of Computational Physics},
  volume={428},
  pages={109985},
  year={2021},
  publisher={Elsevier}
}

@article{chen2021physics,
  title={Physics-informed learning of governing equations from scarce data},
  author={Chen, Zhao and Liu, Yang and Sun, Hao},
  journal={Nature communications},
  volume={12},
  number={1},
  pages={1--13},
  year={2021},
  publisher={Nature Publishing Group}
}

@book{rudin1976principles,
  title={Principles of mathematical analysis},
  author={Rudin, Walter and others},
  volume={3},
  year={1976},
  publisher={McGraw-hill New York}
}

@article{oprea2023learning,
  title={Learning the Delay Using Neural Delay Differential Equations},
  author={Oprea, Maria and Walth, Mark and Stephany, Robert and Nothaft, Gabriella Torres and Rodriguez-Gonzalez, Arnaldo and Clark, William},
  journal={arXiv preprint arXiv:2304.01329},
  year={2023}
}

@article{messenger2022asymptotic,
  title={Asymptotic consistency of the WSINDy algorithm in the limit of continuum data},
  author={Messenger, Daniel A and Bortz, David M},
  journal={arXiv preprint arXiv:2211.16000},
  year={2022}
}

@book{reddy2013introductory,
  title={Introductory functional analysis: with applications to boundary value problems and finite elements},
  author={Reddy, B Dayanand},
  volume={27},
  year={2013},
  publisher={Springer Science \& Business Media}
}

@book{langtangen2003computational,
  title={Computational partial differential equations: numerical methods and diffpack programming},
  author={Langtangen, Hans Petter},
  volume={2},
  year={2003},
  publisher={Springer Berlin}
}

@book{evans2022partial,
  title={Partial differential equations},
  author={Evans, Lawrence C},
  volume={19},
  year={2022},
  publisher={American Mathematical Society}
}

@book{liu2003computational,
  title={Computational inverse techniques in nondestructive evaluation},
  author={Liu, Gui-Rong and Han, Xu},
  year={2003},
  publisher={CRC press}
}

@book{kreyszig1991introductory,
  title={Introductory functional analysis with applications},
  author={Kreyszig, Erwin},
  volume={17},
  year={1991},
  publisher={John Wiley \& Sons}
}

\appendix
\section{The Master Weight Function}
\label{Appendix:Master_Weight_Function}

In this appendix, we show that given any pair of weight functions --- $w$ and $\tilde{w}$ --- we can express $\tilde{w}$ as the composition of $w$ with an affine map.
We then use this result to motivate the \emph{master weight function}.

\bigskip 

To begin, recall that a weight function, $w$, with center $(t_0, X_0)$ and radius $r > 0$ is defined as follows:

$$
w(t, X) = 
\begin{cases} \exp\left( \frac{\beta r^2}{(t - t_0)^2 - r^2} + \beta \right) \prod_{i = 1}^{d} \exp\left(\frac{\beta r^2}{([X]_i - [X_0]_{i})^2 - r^2} + \beta\right) & \text{if } (t, X) \in B^{\infty}_{r}\left(t_0, X_0\right) \\
0 & \text{otherwise} 
\end{cases}
$$

Suppose we have a second weight function, $\tilde{w}$, which has center $(\tilde{t_0}, \tilde{X_0})$ and radius $\tilde{r}$.
Then, 

\begin{align*}
\tilde{w}\left((t - t_0)\frac{\tilde{r}}{r} + \tilde{t}_0, (X - X_0)\frac{\tilde{r}}{r} + \tilde{X}_0\right) &= \exp\left( \frac{\beta}{\left( \frac{(t - t_0)\tilde{r}/r}{\tilde{r}} \right)^2 - 1} + \beta \right) \prod_{i = 0}^{d} \exp\left(\frac{\beta}{\left(\frac{([X]_i - [X_0]_i)(\tilde{r}/r)}{\tilde{r}}\right)^2 - 1} + \beta\right) \\
&= \exp\left( \frac{\beta}{\left( \frac{t - t_0}{r} \right)^2 - 1} + \beta \right)\prod_{i = 1}^{M} \exp\left(\frac{\beta}{\left(\frac{[X]_i - [X_0]_i}{r}\right)^2 - 1} + \beta\right) \\
&= w(t, X)
\end{align*}

Analogously,

$$w\left((t - \tilde{t}_0)\frac{r}{\tilde{r}} + t_0, \left(X - \tilde{X}_0\right)\frac{r}{\tilde{r}} + X_0 \right) = \tilde{w}(t, X)$$

In particular, this means that

\begin{align}
\left( \frac{d}{dt}\right) \tilde{w}(t, X)    &= \left(\frac{r}{\tilde{r}}\right) \left( \frac{d}{dt} \right) w\left(\left(X - \tilde{X}_0\right)\frac{r}{\tilde{r}} + X_0 \right) \label{Eq:w_tilde:t_Derivative} \\
\left( \frac{d}{dX_i} \right) \tilde{w}(t, X)   &= \left(\frac{r}{\tilde{r}}\right) \left( \frac{d}{dX_i} \right) w\left(\left(X - \tilde{X}_0\right)\frac{r}{\tilde{r}} + X_0 \right) \label{Eq:w_tilde:X_Derivative}
\end{align}

These results allow us to relate the partial derivatives of $\tilde{w}$ to those of $w$. 
In particular, if we already know the derivatives of $w$ and the points in $[0, T] \times \Omega$ at which we need to evaluate $\tilde{w}$ to compute the integrals in \eqref{Eq:Loss:Weak}, then we can determine the necessary derivatives of $\tilde{w}$ without directly differentiating $\tilde{w}$.

\bigskip

\texttt{Weak-PDE-LEARN} uses this result to reduce the number of computations required to initialize a new weight function.
Before training begins, we define a \emph{master weight function} by sampling a point, $(t_M, X_M) \in \text{int}\left([0, T] \times \Omega \right)$, from the interior of the problem domain and then selecting a radius, $r_M > 0$, such the ball $B_r^\infty(t_M, X_M)$ lies in the interior of the problem domain; that is, 

$$B_{r_M}^\infty(t_M, X_M) \subseteq \text{int} \left( [0, T] \times \Omega \right).$$

We then define a quadrature grid on the master weight function's support and evaluate the master weight function and its partial derivatives on this grid.
Whenever we create a new weight function, $\tilde{w}$, with center $(\tilde{t}, \tilde{X})$ and radius $\tilde{r} > 0$, we define the quadrature points for that weight function by applying the following transformation to the quadrature points of the master weight function: 

$$(t, X) \to \left( (t - \tilde{t})\frac{r_M}{\tilde{r}} + t_M, (X - \tilde{X})\frac{r_M}{\tilde{r}} + X_M \right)$$

We can then evaluate the partial derivatives of $\tilde{w}$ at its quadrature points using equations \eqref{Eq:w_tilde:t_Derivative} and \eqref{Eq:w_tilde:X_Derivative}. 
This approach dramatically reduces the number of computations required to evaluate the integrals in equation \eqref{Eq:Loss:Weak} for a new weight function.
Since computing the weak form loss represents a sizable portion of the computations in \texttt{weak-PDE-LEARN}, and because \texttt{weak-PDE-LEARN} frequently creates new random weight functions, this approach significantly reduces \texttt{weak-PDE-LEARN}'s run time.

\end{document}